\pdfoutput=1
\documentclass{article}

\usepackage{microtype}
\usepackage{graphicx}
\usepackage{subfigure}
\usepackage{booktabs}
\usepackage{cite}
\usepackage{amsmath,amssymb,amsfonts}
\usepackage[dvipsnames,table]{xcolor}
\usepackage{fancyhdr}

\usepackage{xspace}
\definecolor{darkgreen}{RGB}{0,180,0}
\usepackage[htt]{hyphenat}
\usepackage{svg}

\usepackage{comment}

\usepackage{hyperref}
\usepackage[noabbrev]{cleveref}
\usepackage{booktabs}
\usepackage{tcolorbox}
\usepackage{listings}
\usepackage{paralist}
\usepackage[color=notecolor]{todonotes}
\definecolor{notecolor}{HTML}{FFF8DC}
\usepackage{algorithm}
\usepackage{algorithmic}

\usepackage{url}
\usepackage{amsmath}
\usepackage{array}
\usepackage{xspace}
\usepackage{multicol}
\usepackage{array}
\usepackage{enumitem}
\usepackage{subcaption}
\usepackage{multirow}
\usepackage{mathtools}

\crefname{lstlisting}{listing}{listings}
\Crefname{lstlisting}{Listing}{Listings}

\definecolor{dkgreen}{rgb}{0,0.6,0}
\definecolor{dkred}{rgb}{0.6,0,0}
\definecolor{codegreen}{rgb}{0,0.6,0}
\definecolor{codegray}{rgb}{0.5,0.5,0.5}
\definecolor{codepurple}{rgb}{0.58,0,0.82}
\definecolor{backcolour}{rgb}{0.95,0.95,0.92}
\definecolor{purple}{RGB}{128,0,128}
\definecolor{indigo}{RGB}{75,0,130}
\definecolor{royalblue}{RGB}{65,105,225}
\definecolor{navy}{RGB}{0,0,128}
\definecolor{codebrown}{rgb}{0.6,0.6,0}

\lstdefinestyle{PyStyle}{
  language=Python,
  basicstyle=\ttfamily\scriptsize, 
  aboveskip = 0.05in,
  belowskip = 0.05in,
  breaklines=true,
  float=tp,
  floatplacement=tbp,
  frame=none,
  numbers=none,
  keepspaces=true,
  captionpos=b,
  showstringspaces=false,
  emph={__init__},
  stringstyle=\color{dkgreen},
  emphstyle=\ttb\color{dkred},
  keywordstyle=\color{blue},
  commentstyle=\color{codegreen},
  morekeywords={self,def, for, sum, in, and}
}

\newcommand{\PP}[1]{
\vspace{2px}
\noindent{\bf\textsc{#1}.}\xspace
}

\newcommand{\PPP}[1]{
\vspace{0.05in}
\noindent{\textit{\IfEndWith{#1}{.}{#1}{#1.}}}
}

\newcommand{\squishitemize}{
 \begin{list}{$\bullet$}
  { \setlength{\itemsep}{0pt}
     \setlength{\parsep}{0pt}
     \setlength{\topsep}{0pt}
     \setlength{\partopsep}{0pt}
     \setlength{\leftmargin}{1.95em}
     \setlength{\labelwidth}{1.5em}
     \setlength{\labelsep}{0.5em} } }

\newcounter{Lcount}
\newcommand{\squishlist}{
    \begin{list}{\arabic{Lcount}. }
   { \usecounter{Lcount}
        \setlength{\itemsep}{0pt}
        \setlength{\parsep}{3pt}
        \setlength{\topsep}{0pt}
        \setlength{\partopsep}{0pt}
        \setlength{\leftmargin}{2em}
        \setlength{\labelwidth}{1.5em}
        \setlength{\labelsep}{0.5em} } }

\newcommand{\squishend}{\end{list}}

\newcommand{\bit}{\begin{compactitem}}
\newcommand{\eit}{\end{compactitem}}
\newcommand{\ben}{\begin{compactenum}}
\newcommand{\een}{\end{compactenum}}

\newcommand{\eg}{\textit{e}.\textit{g}.,\xspace}

\newcommand{\cl}[1]{\textcolor{teal}{\textbf{CL: } #1}}
\newcommand{\kr}[1]{\textbf{\color{blue}{KR:#1}}\unskip\color{black}}
\newcommand{\rb}[1]{\textbf{\color{red}{RB:#1}}\unskip\color{black}}
\newcommand{\dm}[1]{\textbf{\color{darkgreen}{AG:#1}}\unskip\color{black}}
\newcommand{\rs}[1]{\textbf{\color{purple}{RS:#1}}\unskip\color{black}}

\newcommand{\sys}{\mbox{\textsc{Stream2LLM}}\xspace}

\newcommand{\minihead}[1]{{\vspace{.4em}\noindent\textbf{#1.}}}

\newcommand{\notice}[1]{}
\renewcommand{\kr}[1]{}
\renewcommand{\rb}[1]{}
\renewcommand{\dm}[1]{}
\renewcommand{\rs}[1]{}
\renewcommand{\cl}[1]{}

\usepackage{hyperref}

\AtBeginDocument{%
  \setlength{\marginparsep}{11pt}%
  \setlength{\topmargin}{-24.95781pt}%
  \setlength{\marginparwidth}{65pt}%
  \setlength{\marginparpush}{5pt}%
}

\usepackage[accepted]{mlsys2026}

\makeatletter
\@ifundefined{isaccepted}{%
  \gdef\isaccepted{1}%
  \@ifundefined{mlsys@appearing}{}{\renewcommand{\Notice@String}{\mlsys@appearing}}%
}{}
\makeatother

\newcommand{\acmbadgeurl}{https://www.acm.org/publications/policies/artifact-review-and-badging-current}
\fancypagestyle{firstpage}{%
  \fancyhf{}%
  \renewcommand{\headrulewidth}{0pt}%
  \fancyhead[R]{\smash{\raisebox{-10pt}{%
    \href{\acmbadgeurl}{\includegraphics[width=55pt]{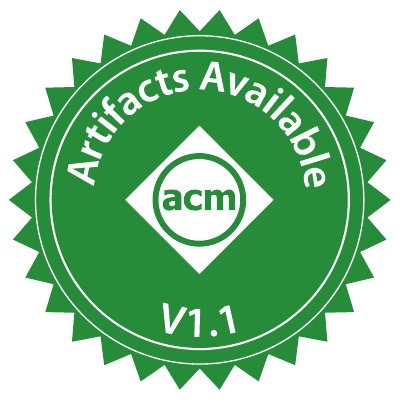}}%
    \hspace{2pt}%
    \href{\acmbadgeurl}{\includegraphics[width=55pt]{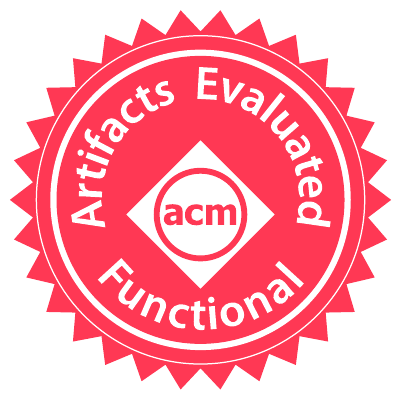}}%
    \hspace{2pt}%
    \href{\acmbadgeurl}{\includegraphics[width=55pt]{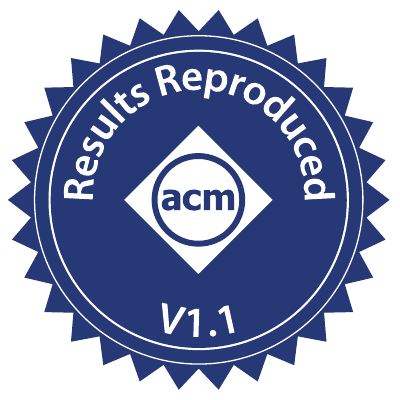}}%
  }}}%
}

\makeatletter
\renewcommand{\mlsyscorrespondingauthor}[2]{%
\ifdefined\isaccepted
  \ifdefined\mlsyscorrespondingauthor@text
    \g@addto@macro\mlsyscorrespondingauthor@text{, #1 #2}%
  \else
    \gdef\mlsyscorrespondingauthor@text{#1 #2}%
  \fi
\fi
}
\makeatother

\mlsystitlerunning{Stream2LLM: Overlap Context Streaming and Prefill for Reduced TTFT}

\begin{document}

\twocolumn[
\mlsystitle{Stream2LLM: Overlap Context Streaming and Prefill for Reduced Time-to-First-Token}

\mlsyssetsymbol{equal}{*}

\begin{mlsysauthorlist}
\mlsysauthor{Rajveer Bachkaniwala}{GT}
\mlsysauthor{Chengqi Luo}{GT}
\mlsysauthor{Richard So}{GT}
\mlsysauthor{Divya Mahajan}{GT}
\mlsysauthor{Kexin Rong}{GT}
\end{mlsysauthorlist}

\mlsysaffiliation{GT}{Georgia Tech, Atlanta, USA}

\mlsyscorrespondingauthor{Rajveer Bachkaniwala}{\href{mailto:rr@gatech.edu}{rr@gatech.edu}, \href{https://x.com/rajveerbach}{@rajveerbach on X}}

\mlsyskeywords{LLM inference, streaming, KV cache, time-to-first-token, request scheduling, prefill}

\begin{abstract}

Context retrieval systems for LLM inference face a critical challenge: high retrieval latency creates a fundamental tension between waiting for complete context (poor time-to-first-token) and proceeding without it (reduced quality). Streaming context incrementally--overlapping retrieval with inference--can mitigate this latency, but doing so with concurrent requests introduces new challenges: requests contend for GPU compute and memory, and scheduling must adapt to dynamic context arrivals.

We present \sys, a streaming-aware LLM serving system for concurrent prefill-decode disaggregated deployments.
\sys introduces adaptive scheduling and preemption for two distinct retrieval patterns: \emph{append-mode} (progressive context accumulation) and \emph{update-mode} (iterative refinement with cache invalidation).
It decouples scheduling decisions from resource acquisition, enabling flexible preemption strategies guided by hardware-specific cost models, and uses longest common prefix matching to minimize redundant computation when input changes dynamically.
To evaluate \sys, we collect two large-scale, real-world streaming workloads based on web crawling and approximate nearest neighbor search.
Our evaluation demonstrates that streaming architecture delivers up to 11$\times$ TTFT improvements, with cost-aware scheduling providing critical benefits under memory pressure, all while maintaining throughput parity with non-streaming baselines.

Code: \href{https://github.com/rajveerb/stream2llm/tree/mlsys_artifact}{\texttt{https://github.com/rajveerb/stream2llm/}}

\end{abstract}

\vskip 0.3in
]

\printAffiliationsAndNotice{}
\thispagestyle{firstpage}

\suppressfloats[t]
\begin{figure}[t]
    \centering
    \includegraphics[width=0.9\columnwidth]{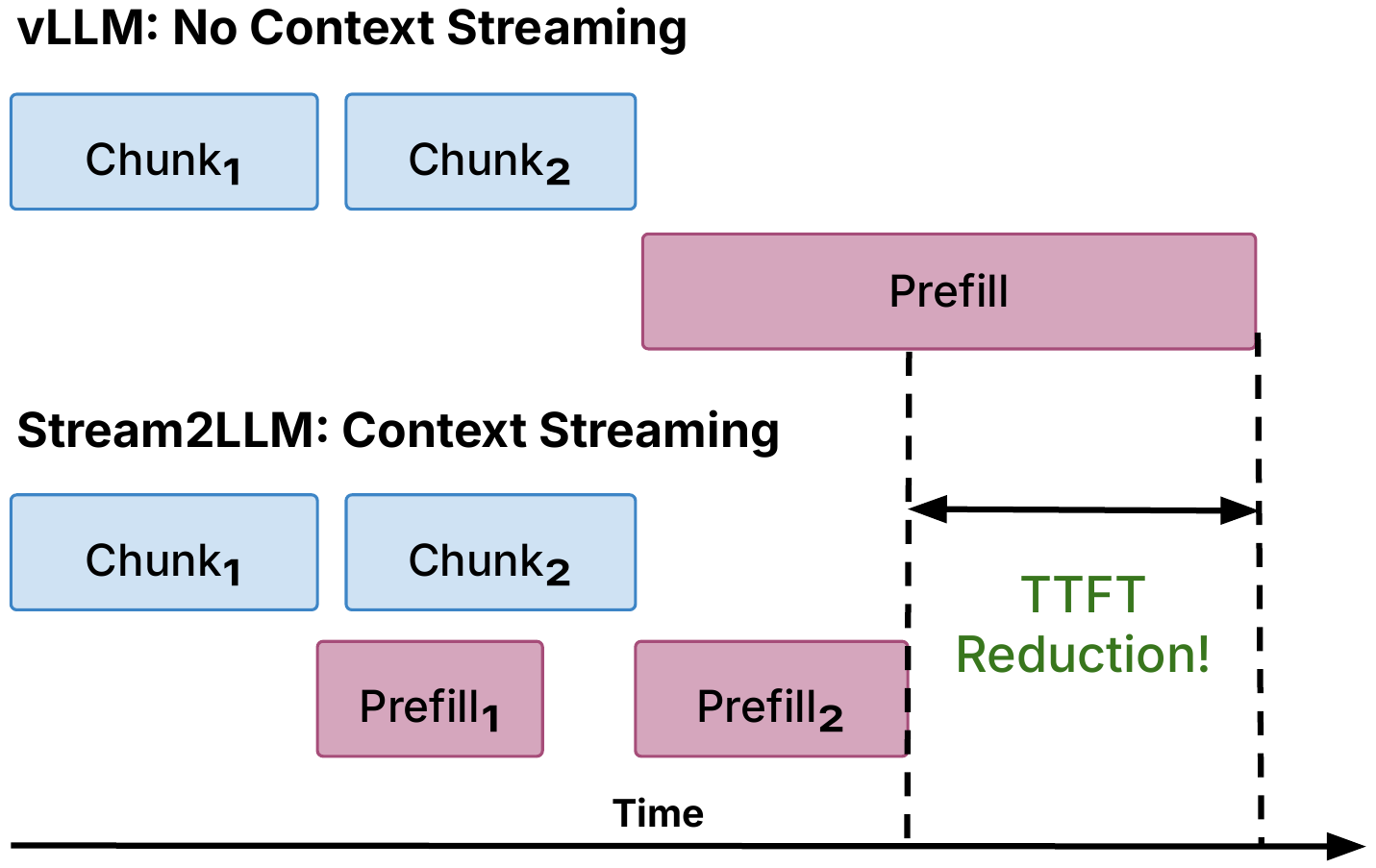}
    \vspace{-1.5em}
    \caption{Context streaming overlaps retrieval with prefill, reducing TTFT by beginning inference as chunks arrive.}
    \label{fig:streaming-retrieval-prefill}
    \vspace{-1.5em}
\end{figure}

\section{Introduction}
\label{sec:introduction}

Modern large language models (LLMs) increasingly rely on external context retrieval to provide accurate, up-to-date responses~\cite{lewis2020retrieval,gao2023retrieval,xu2023retrieval}. Context retrieval mechanisms, such as fetching relevant pages from web search or vector search (ANNS) over large vector databases, can take hundreds of milliseconds to seconds. This creates a fundamental tension between two undesirable outcomes: 
the system can either wait for all context to arrive, which increases time-to-first-token and degrades user interactivity, or begin generation early with partial context, which can reduce response quality.

A natural approach is to stream context incrementally, feeding chunks to the model as they arrive rather than waiting for retrieval to complete (\autoref{fig:streaming-retrieval-prefill}). In single-request settings, this can significantly reduce latency~\cite{jiang2024piperag,yu2025aquapipe}. Production deployments, however, must balance per-user interactivity against serving cost: batching concurrent requests onto shared GPU compute and memory improves throughput and profit margins but directly increases time-to-first-token (key metric for responsiveness). Moreover, context arrives asynchronously and at varying rates across requests, requiring the system to dynamically adapt scheduling while managing competition for shared GPU resources.

\begin{figure}[t]
  \centering
  \includegraphics[width=\columnwidth]{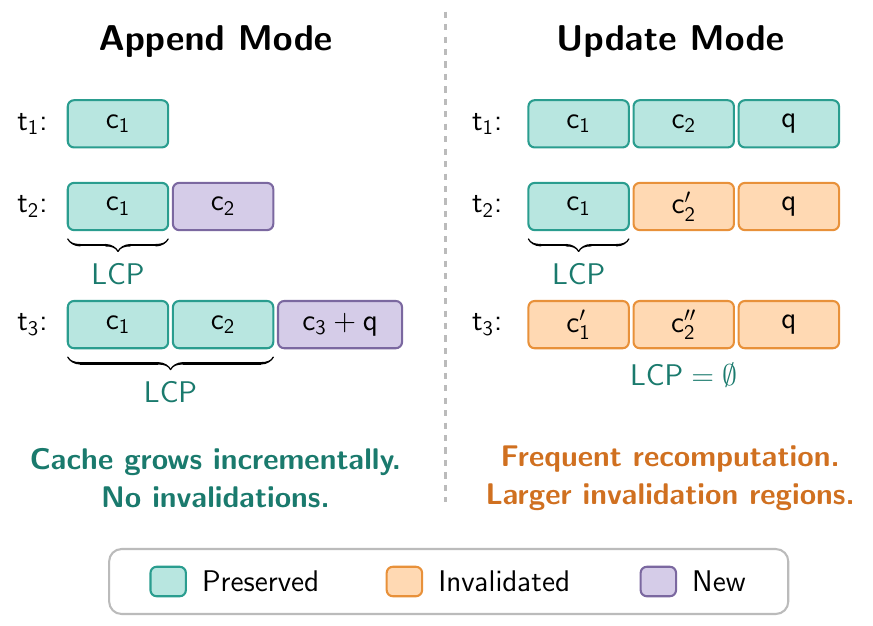}
  \vspace{-1.5em}
  \caption{\sys supports two retrieval patterns--append-mode and update-mode--and uses longest common prefix (LCP) matching to minimize cache invalidation by preserving shared prefixes across input updates.}
  \label{fig:streaming-modes-lcp}
  \vspace{-1.5em}
\end{figure}

Streaming with concurrent requests introduces three interrelated challenges. First, GPU memory becomes contested: each request maintains KV cache blocks for its growing input, and memory exhaustion forces preemption---either swapping cache to CPU or discarding it for later recomputation. Second, the scheduler must decide which requests to prioritize as context chunks arrive at different times, balancing responsiveness, fairness, and cache locality. Third, context retrieval itself exhibits two distinct patterns (\autoref{fig:streaming-modes-lcp}): \emph{append-mode}, where chunks progressively extend the input, and \emph{update-mode}, where the context set changes iteratively as search algorithms refine results. Na\"ive approaches that treat all streaming requests uniformly suffer from poor scheduling or excessive cache recomputation.

We present \sys, a system that extends the vLLM inference engine~\cite{kwon2023efficient} to support streaming inputs with concurrent requests.
\sys targets the prefill stage in a prefill-decode disaggregated deployment~\cite{zhong2024distserve,patel2023splitwise,qin2024mooncake}, following industry practice. 
This architecture enables optimization for time-to-first-token (TTFT) and throughput, where streaming decisions matter most, without impacting decode-stage latency, i.e., time-per-output-token (TPOT), which is handled by separate decode instances.

Specifically, \sys introduces a two-phase scheduling architecture that decouples priority-based request ordering from resource acquisition.
This separation of concerns--deciding what to run versus how to allocate resources--enables flexible preemption strategies without hardcoded policies.
When memory is exhausted, the system preempts low-priority requests using cost-based decisions guided by hardware-specific performance models, choosing between recomputing invalidated cache and swapping cache blocks to CPU. 
To minimize redundant computation when inputs change dynamically, \sys uses longest common prefix (LCP)-based cache invalidation, preserving cache for unchanged prefix tokens while invalidating only changed portions. 
This mechanism uses a single cache management strategy for both append-mode and update-mode retrieval patterns.

We conduct a comprehensive evaluation using real-world streaming traces collected from web crawling and ANNS-based retrieval systems, capturing both append-mode and update-mode retrieval patterns.
Experiments on H100 and H200 GPUs with Llama-3.1 models show that \emph{while streaming architecture provides substantial benefits across different schedulers, intelligent scheduling becomes increasingly important as load increases}. At low loads, all streaming schedulers converge to similar performance, improving time-to-first-token by up to 3.9-11.0$\times$ over non-streaming baselines. 
However, as concurrency and memory contention increase, scheduler choice becomes critical: proper request prioritization strategies enable efficient cache reuse and maintain responsiveness, while na\"ive scheduling causes catastrophic tail latency (up to 10$\times$ worse under extreme memory pressure). These findings hold across GPU architectures and workload types, demonstrating that streaming is necessary but insufficient; intelligent scheduling and cache management is essential for making streaming viable in production deployments.

In summary, this paper makes the following contributions:

\squishitemize
\item We present \sys, a system that extends vLLM to support concurrent streaming inputs in LLM inference, handling both \emph{append-mode} (progressive context accumulation) and \emph{update-mode} (iterative context refinement) workloads.
\item We introduce a two-phase scheduling architecture that decouples request prioritization from GPU memory allocation and preemption, enabling flexible policies that improve KV cache reuse. This is combined with cost-based adaptive preemption (recomputation vs. swapping) and longest common prefix-based cache invalidation to reduce redundant computation for dynamically changing inputs.
\item We evaluate \sys using real-world, streaming traces from web crawling and ANNS-based retrieval systems. Results show that streaming with intelligent scheduling and cache management delivers up to $11\times$ improvements in TTFT.

\squishend

\section{Background}
\label{sec:background}

\subsection{LLM Inference and KV Cache Management}

Large language model inference follows an autoregressive generation process consisting of two distinct phases. Given an input with $n$ tokens $\mathbf{x} = [x_1, x_2, \ldots, x_n]$, the \emph{prefill phase} processes all input tokens in parallel, computing attention scores and generating key-value pairs $(K_i, V_i)$ for each token position $i \in [1, n]$. These KV pairs are stored in GPU memory as the KV cache. The \emph{decode phase} then generates output tokens autoregressively: at each step $t$, the model attends to all previously computed KV pairs $\{(K_1, V_1), \ldots, (K_{n+t-1}, V_{n+t-1})\}$ to generate the next token $x_{n+t}$. The newly generated token's KV pair $(K_{n+t}, V_{n+t})$ is appended to the cache, avoiding recomputation of attention for tokens processed in earlier steps.

For a transformer model with $L$ layers, hidden dimension $d$, total attention heads $h$, and $h_{\text{kv}}$ key-value heads (head dimension $d_h = d/h$), each token position stores $2 L d \tfrac{h_{\text{kv}}}{h}$ values across all layers (keys and values). The total KV cache memory for a sequence of length $\ell$ is
$M_{\text{KV}} = 2 L \ell d \tfrac{h_{\text{kv}}}{h} \cdot b$,
where $b$ is the number of bytes per parameter (typically 2 bytes for FP16). For Meta Llama-3.1-8B with $L=32$, $d=4096$, $h=32$, $h_{\text{kv}}=8$, and $\ell=32\text{K}$, this amounts to $M_{\text{KV}} \approx 4.0$\,GB per request in FP16 precision. When serving $B$ concurrent requests through batching, total memory consumption becomes $B \cdot M_{\text{KV}}$. As $B$ increases to improve throughput, available memory per request decreases, forcing systems to either limit batch size or preempt requests to free KV cache.

\subsection{vLLM and PagedAttention}
\label{sec:bg}

Traditional KV cache implementations allocate a single contiguous memory region of size $M_{\text{KV}}$ for each request, leading to memory fragmentation as requests arrive and complete at different times. vLLM~\cite{kwon2023efficient} addresses this through PagedAttention, which divides the KV cache into fixed-size blocks. For a block size of $k$ tokens, the KV cache for a sequence is stored as a sequence of blocks $\mathcal{B} = \{B_1, B_2, \ldots, B_{\lceil \ell/k \rceil}\}$, where each block $B_j$ stores KV pairs for tokens $[(j-1)k + 1, \min(jk, \ell)]$. These blocks need not be contiguous in physical memory, similar to virtual memory paging in operating systems.

The attention computation in PagedAttention operates block-wise. For computing attention at position $t$, the system retrieves all blocks $\{B_1, \ldots, B_{\lceil t/k \rceil}\}$ containing KV pairs for positions $[1, t]$, regardless of their physical memory locations. This design provides two key benefits: (1) reduced fragmentation by allocating blocks on-demand from a free block pool, and (2) efficient prefix sharing when multiple requests share common input prefixes, as blocks containing the shared prefix can be referenced by multiple requests without duplication~\cite{gim2023prompt}.

vLLM supports continuous batching~\cite{yu2022orca}, maintaining a dynamic batch $\mathcal{R}_t = \{r_1, r_2, \ldots, r_{B_t}\}$ where requests can be added or removed at any scheduling step $t$. When total KV cache memory exceeds GPU capacity $M_{\text{GPU}}$, the system must preempt requests. For a request $r$ with $\ell_r$ processed tokens occupying $\lceil \ell_r / k \rceil$ blocks, preemption follows one of two strategies:
\squishitemize
\item \textbf{Recomputation}: Discard all KV cache blocks for $r$, freeing $\lceil \ell_r / k \rceil \cdot M_{\text{block}}$ memory (where $M_{\text{block}} = 2 L k d \tfrac{h_{\text{kv}}}{h} \cdot b$). Upon resumption, recompute the prefill phase for all $\ell_r$ tokens, incurring compute cost $C_{\text{recomp}}(r) = \ell_r \cdot C_{\text{prefill}}$ where $C_{\text{prefill}}$ is the per-token prefill latency.
\item \textbf{Swapping}: Transfer all blocks $\{B_1, \ldots, B_{\lceil \ell_r / k \rceil}\}$ from GPU to CPU memory, incurring data transfer cost $C_{\text{swap}}(r) = \frac{\lceil \ell_r / k \rceil \cdot M_{\text{block}}}{BW_{\text{PCIe}}}$ where $BW_{\text{PCIe}}$ is the PCIe bandwidth. Upon resumption, swap blocks back to GPU with symmetric cost. The blocks preserve computed KV values, avoiding recomputation.
\squishend

Selecting between the two strategies requires comparing $C_{\text{recomp}}(r)$ versus $2 \cdot C_{\text{swap}}(r)$ (accounting for bidirectional transfer); \sys uses this cost model to guide its adaptive preemption decisions, as described in \S\ref{sec:preemption-mgmt}.

\subsection{Context Retrieval for LLMs}

Modern LLM deployments require external context $\mathcal{C}$ to augment the model's parametric knowledge~\cite{lewis2020retrieval}. Given a query $q$, context retrieval systems produce a set of relevant documents $\mathcal{D} = \{d_1, d_2, \ldots, d_m\}$ that are concatenated with $q$ to form the final input. Two retrieval mechanisms are commonly used:

\textbf{Web crawler retrieval}: Given a query $q$, the system fetches documents from the web or internal knowledge bases. Latency $T_{\text{web}}$ includes network round-trip time, DNS resolution, content fetching, and parsing, typically ranging from 200ms to multiple seconds per document.

\textbf{ANNS-based retrieval}: Given a query embedding $\mathbf{e}_q \in \mathbb{R}^{d_e}$ and a corpus $\mathcal{C} = \{c_1, \ldots, c_N\}$ with embeddings $\{\mathbf{e}_{c_1}, \ldots, \mathbf{e}_{c_N}\}$, find the $k$-nearest neighbors
$\mathcal{D} = \text{top-}k(\{c_i : \text{sim}(\mathbf{e}_q, \mathbf{e}_{c_i}) \mid c_i \in \mathcal{C}\})$, 
where $\text{sim}(\cdot, \cdot)$ is a similarity metric (\eg cosine similarity or L2 distance). 
While in-memory ANNS solutions such as HNSW~\cite{malkov2018efficient} achieve low retrieval latency, storing the full index in memory is infeasible at scale ($N \sim 10^9$).
Disk-based ANNS solutions such as DiskANN~\cite{jayaram2019diskann} and SPANN~\cite{chen2021spann} address this by storing indexes on disk, where search performs $O(\log N)$ graph traversals with disk I/O at each step.
They are widely deployed in production due to cost: SSDs are orders of magnitude cheaper per capacity than DRAM~\cite{AISAQ,diskann-gtc}.
End-to-end retrieval latency $T_{\text{ANNS}}$ ranges from 100ms to several seconds depending on corpus size, index complexity parameter (e.g., search list size in DiskANN), disk bandwidth, and pipeline overhead (query embedding, network, post-processing).

\subsection{Traditional vs Streaming Inference}
Traditional inference systems construct the complete input $\mathbf{x} = [\mathcal{D}, q]$ only after retrieval finishes at time $T_{\text{retrieve}}$, then begin the prefill phase. This results in time-to-first-token (TTFT) of $T_{\text{TTFT}} = T_{\text{retrieve}} + T_{\text{prefill}}(|\mathcal{D}| + |q|)$, where $T_{\text{retrieve}}$ often dominates.
Streaming context approaches reduce TTFT by sending context chunks incrementally. Instead of waiting for complete $\mathcal{D}$, documents arrive over time: $d_1$ at time $t_1$, $d_2$ at time $t_2$, etc. The LLM begins inference with partial context $\mathcal{D}_t = \{d_i : t_i \leq t\}$ and continues as new documents arrive. 

We consider two complementary streaming modes that capture distinct classes of retrieval workloads. 
In \emph{append mode}, context grows monotonically ($\mathcal{D}_t \subseteq \mathcal{D}_{t'}$ for $t < t'$), corresponding to pipelines that produce results incrementally without revisiting earlier results. 
For example, in web crawling pipelines, documents arrive as pages are retrieved and can be streamed immediately when no global post-processing (e.g., reranking) is required.
More broadly, append mode applies whenever documents are emitted in final form upon arrival.
In \emph{update mode}, the context is refined over time, with $\mathcal{D}_t$ evolving as higher-quality candidates replace earlier ones, corresponding to pipelines that progressively refine retrieval results.
For example, AquaPipe~\cite{yu2025aquapipe} enables early return of partial ANNS results before search completion; Vespa implements phased ranking where documents are progressively filtered and re-ranked across stages~\cite{vespa2024rag}; and distributed search engines such as Elasticsearch similarly emit partial results as individual shards complete. 
These pipelines naturally produce a sequence of refined top-$k$ sets, which can be streamed to the inference system.

These modes have been explored separately in prior systems—PipeRAG~\cite{jiang2024piperag} for append and AquaPipe~\cite{yu2025aquapipe} for update—and have been evaluated in single-request settings ($B=1$). 
In this work, we support both modes under a unified concurrent setting.

\section{Challenges}
\label{sec:challenges}

Streaming concurrent inference workloads differ fundamentally from static batch inference.
Requests stream context incrementally over time, evolve asynchronously, and compete for shared GPU resources.
These unique properties create several system design challenges.

\minihead{Dynamic Workload Variability}
Each request expands dynamically as new chunks of context arrive and generates outputs of unpredictable length until termination tokens appear.
This variability makes both compute demand and memory footprint time-dependent, and requires dynamic policies.

Chunk arrivals are asynchronous: some requests stall waiting for retrieval while others suddenly receive new context.
Requests that just received a chunk should be prioritized to process it while the request's KV cache blocks remain allocated in GPU memory, but traditional scheduling policies (\eg FCFS or progress-based heuristics) cannot capture this temporal priority.
Schedulers must dynamically re-rank requests as chunk arrival patterns shift, balancing freshness, fairness, and preemption cost.

All requests draw from the same finite pool of GPU memory for KV cache blocks.
As input sequences grow, total cache usage can exceed GPU capacity, forcing the system to free memory through preemption.
The optimal strategy--whether to recompute discarded cache or to swap it to CPU memory--depends on request progress, hardware bandwidth ratios, and expected resumption time.

\minihead{Mode-Dependent Context Evolution}
Streaming retrieval workloads differ in how input sequences evolve. In web-crawler-style retrievers (append mode), requests extend the existing sequences, whereas in ANNS-style retrievers (update mode), requests replace parts of the sequence as retrieval refines results.
These modes require different scheduling and cache management behaviors.

In update mode, the system must determine which KV cache blocks remain valid after each update.
If the scheduler naively invalidates all blocks, it wastes memory by discarding valid cached results and forces expensive recomputation. Conversely, if the scheduler reuses blocks without verification, it risks computing incorrect output based on stale cache, since subsequent attention computations would attend to outdated key-value pairs that no longer correspond to the current input sequence.
Frequent replacements early in the sequence can lead to high recomputation costs.

Append and update workloads also differ in optimal preemption strategies. Append-mode requests favor swapping to preserve reusable cache, while update-mode requests often prefer recomputation to avoid retaining soon-invalid data.

\section{System Design}
\label{sec:overview}

\minihead{Deployment Assumption} \sys targets the prefill instance in a prefill-decode disaggregated architecture~\cite{qin2024mooncake}, where prefill and decode run on separate GPU pools. This is standard practice in production LLM serving~\cite{zhong2024distserve,patel2023splitwise}. TTFT and throughput are the metrics relevant to prefill instances; token generation latency (TPOT) is handled by decode instances and is not evaluated.

\sys addresses the challenges with concurrent, streaming workloads with two key design principles:

\squishitemize
    \item \textbf{Support for two distinct streaming modes:} Requests can either append new input chunks to the existing sequence or replace parts of the input sequence entirely. Each mode imposes different requirements on cache management, scheduling, and preemption.
    \item \textbf{Decoupling scheduling from resource acquisition:} \sys separates the decision of which request to prioritize from the mechanics of allocating GPU memory and applying preemption. This decoupling enables sophisticated policies that maximize KV cache reuse while adapting to dynamic workloads.
\squishend

The subsequent subsections introduce the two-phase scheduling design (\S~\ref{sec:scheduling-state-design}), cache invalidation mechanisms (\S~\ref{sec:kv-cache-invalidation}), cost-based preemption strategies (\S~\ref{sec:preemption-mgmt}), and scheduling policies (\S~\ref{sec:scheduling-policies}).

\begin{figure}[t]
  \centering
  \includegraphics[width=\columnwidth]{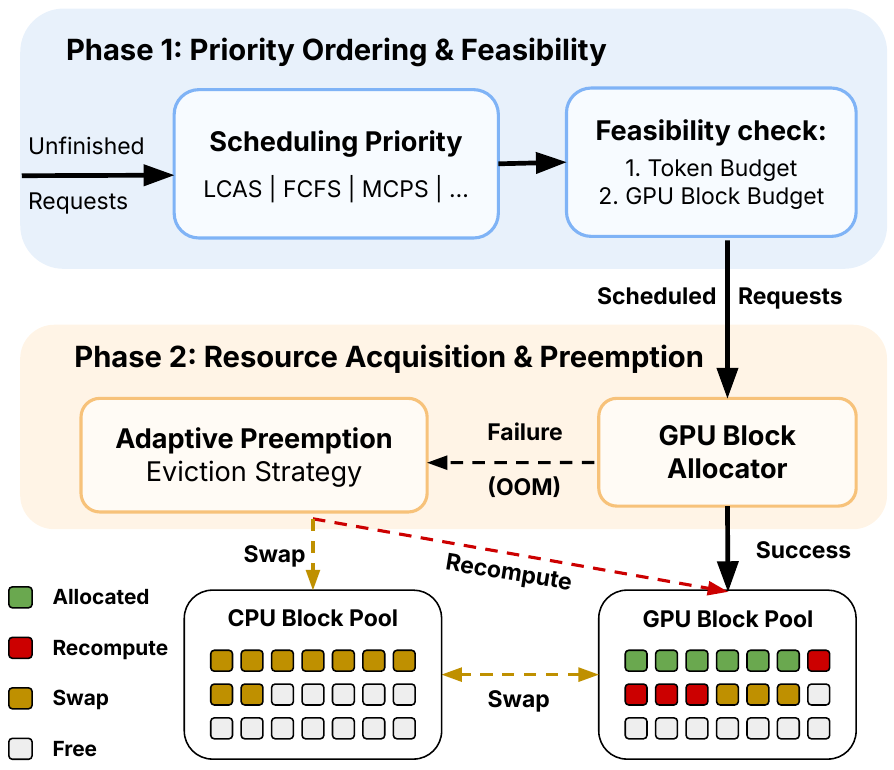}
  \vspace{-2em}
  \caption{Overview of \sys's two-phase scheduler that separates scheduling decisions from resource acquisition. Phase 1 determines request priority and feasibility, while Phase 2 allocates KV cache blocks and applies preemption under memory pressure.}
  \label{fig:sys-design}
\end{figure}

\subsection{Two-Phase Scheduling and Request Lifecycle}
\label{sec:scheduling-state-design}

A key insight in \sys is the separation of \emph{scheduling decisions} from \emph{resource acquisition}. 
Rather than embedding both in a single control loop, \sys organizes them into two distinct phases, allowing scheduling policies to reason about priorities independently of allocation and preemption mechanics.
\autoref{fig:sys-design} illustrates this two-phase architecture.

When scheduling, allocation, and preemption are tightly coupled, several challenges arise:

\squishitemize
  \item \emph{Fixed Policy Coupling.} A monolithic loop forces the scheduler to preempt immediately when resources are exhausted, using a static rule that may contradict the policy’s notion of priority. This prevents different scheduling algorithms from expressing distinct preemption behaviors.

  \item \emph{Cache Invalidation Coupling.} 
  When an input sequence is updated, two things must happen: the token count increases and cached KV blocks may become invalid. In a monolithic loop, these operations interleave with scheduling and allocation, making it unclear whether to invalidate before or after attempting allocation.
\squishend

\sys decouples scheduling and resource acquisition into two phases that explicitly separate concerns:

\textbf{Phase 1: Priority Ordering and Feasibility Analysis.} The scheduler invokes the selected scheduling algorithm (\S~\ref{sec:scheduling-policies}) to compute an ordered list of all unfinished requests ranked by priority. It then performs a feasibility analysis that determines which requests can theoretically be scheduled given the current token budget (i.e., the maximum number of tokens that can be processed in a scheduling step) and estimated GPU block requirements.
Specifically, for each request in priority order, the scheduler estimates the number of KV cache blocks required (based on computed and new tokens) and checks whether sufficient free GPU blocks remain; if not, the request is marked infeasible.
Crucially, this phase performs no resource allocation and modifies no request state---it only computes feasibility. Requests that cannot fit within current resources are added to a \texttt{not\_scheduled\_reqs} list while preserving their priority.

\textbf{Phase 2: Resource Acquisition with Adaptive Preemption.} The scheduler attempts to allocate GPU blocks for each request selected in Phase 1.
    If allocation fails, the scheduler selects a preemption candidate from the \texttt{not\_scheduled\_reqs} list in reverse priority order (i.e., lowest-priority first).
    Because requests in \texttt{not\_scheduled\_reqs} may still hold KV cache blocks from previous scheduling steps, preempting them releases those blocks, freeing GPU memory to proceed with allocation for higher-priority requests.
    The scheduler then applies the decision framework from \S~\ref{sec:preemption-mgmt} to choose between recomputation and swapping as the eviction strategy.

This two-phase structure separates \textit{what} to run from \textit{how} to allocate resources and provides several benefits. 
The separation ensures that priority decisions are made independently of resource constraints, and preemption naturally follows the scheduling priority order without requiring policy-specific rules. 
Cache invalidation occurs at precise boundaries between analysis and allocation. 
The resource-acquisition phase can also flexibly integrate cost-, or latency-aware strategies without altering core scheduling logic. 

\begin{figure}[t]
  \centering
\includegraphics[width=0.9\columnwidth]{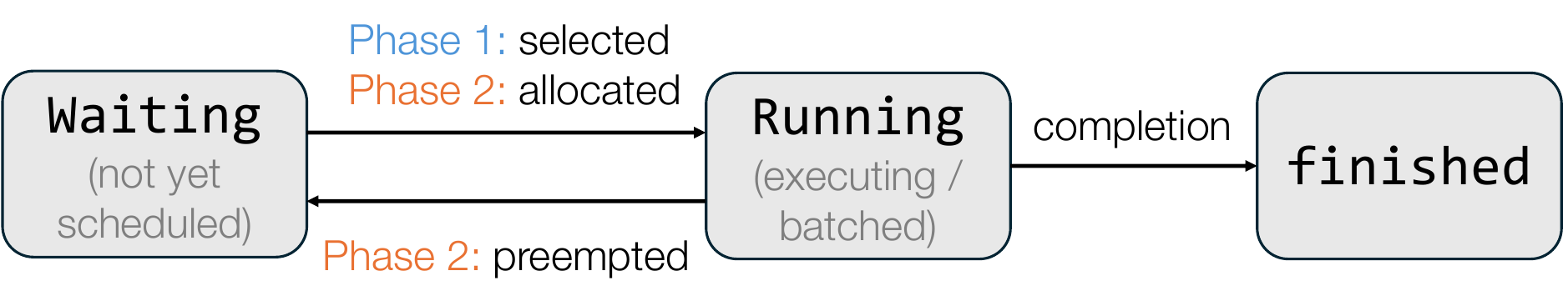}
  \vspace{-1em}
  \caption{\sys request state transitions.}
  \label{fig:state}
  \vspace{-1.5em}
\end{figure}

\minihead{Request Lifecycle} The scheduler tracks requests through three primary queues:
\texttt{waiting} (arrived but not yet scheduled), \texttt{running} (currently in execution or scheduled in the current batch), and \texttt{finished} (completed).
Transitions between these queues are governed by the two-phase design (\autoref{fig:state}). 
Requests move from \texttt{waiting} to \texttt{running} when selected in Phase 1 and allocated in Phase 2. If a running request is preempted, it returns to \texttt{waiting}: recomputation resets its progress, while swapping preserves state by offloading KV blocks to CPU memory. Upon completion, requests transition from \texttt{running} to \texttt{finished}.

\subsection{KV Cache Invalidation for Streaming Inputs}
\label{sec:kv-cache-invalidation}
\sys supports two input sequence modification modes suited to different context retrieval patterns:
\squishitemize
  \item \textbf{Append Mode:} New input chunks are appended to the existing sequences, typical in crawler-style workloads.
  \item \textbf{Update Mode:} The entire input sequence is updated dynamically, typical in ANNS-style workloads.
\squishend

A unique challenge for streaming input workloads is that the input token sequence itself changes dynamically. 
When a new input chunk arrives, the scheduler must decide which cached KV blocks remain valid and which must be invalidated. 
Traditional LLM inference assumes a fixed input at request arrival, computing KV cache blocks once for the entire request lifetime.
However, in streaming context retrieval workloads, the input grows or changes as new document chunks arrive. For example, the input may evolve from $[d_1, q]$ (document chunk 1 + query) to  $[d_1, d_2, q]$ as document chunks are added, or from $[d_1, d_2, q]$ to $[d_1', d_2, q]$ when chunks are replaced in ANNS-style systems. 
If the scheduler naively invalidates all KV cache blocks upon each input update, it wastes memory by discarding previously computed blocks that remain valid and forces expensive recomputation. Conversely, if the scheduler reuses blocks without verification, it risks computing incorrect output based on stale cache.

\minihead{Longest Common Prefix Invalidation}
\sys addresses this by computing the longest common prefix (LCP) between the old and new input token sequences, as illustrated in \autoref{fig:streaming-modes-lcp}.
When a streaming request receives a new input chunk, the engine computes the LCP and invalidates only the KV cache blocks corresponding to tokens \textit{beyond} the LCP. Blocks for tokens within the LCP are preserved, avoiding redundant recomputation.
The LCP approach is optimal when input updates involve appending new chunks or replacing suffix tokens, which is typical in context retrieval systems. It becomes less beneficial when updates are frequent or affect early tokens.

For example, suppose a request previously computed KV cache for tokens $[d_1, d_2, q, \text{output}_1, \text{output}_2]$, and an input update replaces the input with $[d_1, d_2', q, \text{output}_1, \text{output}_2]$, where $d_2' \ne d_2$. The LCP is then $[d_1]$ (length 1). 
The scheduler invalidates cache blocks for tokens 1 onward (corresponding to $d_2, q,$ and output) while preserving the KV cache for token 0 (document chunk 1). 

For analysis, \sys tracks the total number of tokens invalidated across all input updates per request via \texttt{total\_tokens\_invalidated}. This metric is reported when the request finishes, allowing researchers to measure the cost of cache invalidation in context retrieval workloads. High invalidation counts indicate frequent or aggressive input updates, suggesting that the system is spending significant effort on recomputation.

\minihead{Cache Invalidation Semantics} 
When $k$ tokens are invalidated beyond the LCP, the scheduler frees the corresponding KV cache blocks and returns them to the free block pool. 
For CPU-swapped requests, the scheduler also frees the corresponding CPU blocks to reduce memory pressure. 
The scheduler then sets \texttt{num\_computed\_tokens} to the LCP length, forcing recomputation only for tokens that changed.

The interaction between cache invalidation and preemption requires careful sequencing to avoid freeing blocks that are about to be swapped back in.
When a preempted request receives an input update, the scheduler invalidates blocks from the LCP onward on CPU and free these blocks from CPU memory. When the request resumes, it first swaps its remaining blocks (those before the LCP) back to GPU, then recomputes tokens from the LCP onward. 
This design ensures that input updates do not cause memory leaks or inconsistencies in the swapped state.

\begin{figure}[t]
  \centering
  \includegraphics[width=\columnwidth]{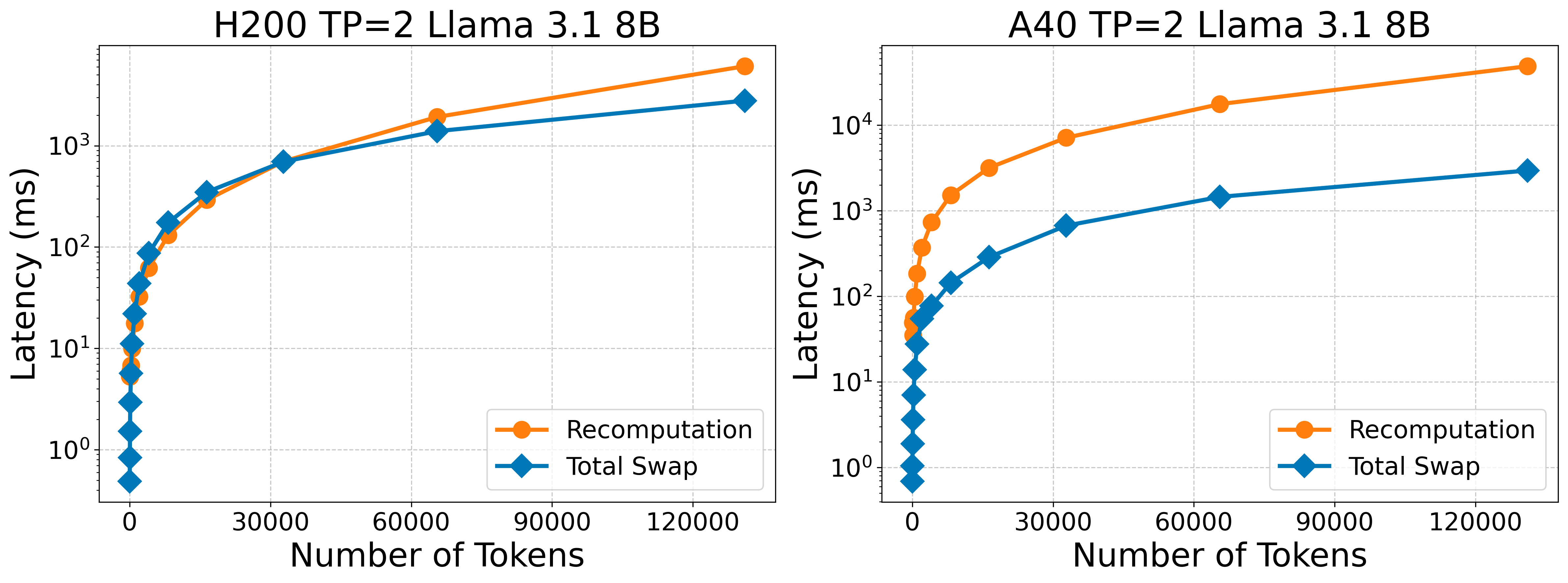}
  \vspace{-1em}
  \caption{Performance models for recomputation vs.\ total swap latency costs across token counts on H200 and A40.}
  \label{fig:perf-model}
  \vspace{-1.5em}
\end{figure}

\begin{table*}[t]
\footnotesize
\centering
\caption{Scheduling algorithm performance on streaming workloads.}
\label{tab:policy}
\begin{tabular}{l|cc|cc}
\toprule
& \multicolumn{2}{c|}{\textbf{Append Mode}} & \multicolumn{2}{c}{\textbf{Update Mode}} \\
\textbf{Policy} & \textbf{Rating} & \textbf{Key Behavior} & \textbf{Rating} & \textbf{Key Behavior} \\
\midrule
DEFAULT vLLM & Poor & Ignores arrival timing & Poor & Ignores recomputation cost \\
FCFS & Good & No recency awareness & Moderate & No update prioritization \\
MCPS & Good & Tracks progress & Poor & Unstable under updates \\
LCAS & Good & Prioritizes recent input & Good & Prioritizes recent updates \\
\bottomrule
\end{tabular}
\end{table*}

\subsection{Cost-based Preemption}
\label{sec:preemption-mgmt}

When GPU memory is exhausted, \sys preempts requests and chooses between recomputation and swapping strategies based on cost models. 
\sys profiles two hardware-specific latency cost functions offline on each target platform, such as shown in \autoref{fig:perf-model}.
$\emph{recomputation\_latency}(T)$ predicts the time to recompute $T$ tokens via prefill. 
We measure prefill latency across varying token counts (1K-128K) on each GPU (A40, H100, H200) and fit a piecewise-linear model to account for memory bandwidth saturation.
$\emph{swap\_latency}(C)$ predicts the time to swap $C$ KV cache blocks between GPU and CPU memory. 
We measure transfer latency for typical block sizes 2 MB (for the given model Llama 3.1 8B, and default block size of 16) and account for PCIe bandwidth constraints. 
On systems with NVLink or faster interconnects (e.g., NVIDIA GB300 and GH200 has up to 450 GB/s host–device bandwidth), swap costs are correspondingly lower.

These cost functions are hardware-specific: a recomputation that takes 50ms on H100 may take 200ms on A40 due to differences in compute density and memory bandwidth. The scheduler evaluates both cost functions at preemption time and selects the strategy with lower predicted latency. The cost functions can be updated dynamically to adapt to changing system characteristics (e.g., contention for PCIe bandwidth), though our current implementation uses static profiles derived from idle-system benchmarks.

\subsection{Scheduling and Eviction Policies}
\label{sec:scheduling-policies}
The scheduling algorithm determines request execution order. Different policies optimize different tradeoffs: throughput, latency, fairness. Each policy selects its lowest-priority request for eviction when GPU memory fills, and uses the cost model from \S~\ref{sec:preemption-mgmt} to decide between recomputation and swapping strategies.

\sys implements multiple scheduling algorithms with different trade-offs, selected via the \texttt{SCHEDULER\_TYPE} variable. \autoref{tab:policy} summarizes each algorithm's performance under different streaming workloads.

\subsubsection{DEFAULT vLLM (variant of FIFO)}
The baseline scheduler uses arrival-time ordering for the waiting queue, processing new requests in FIFO order. For running requests, it maintains their current execution order without explicit arrival-time re-sorting. When GPU memory is exhausted, it preempts the last request in the running queue (LIFO eviction). Preempted requests are re-queued at the front of the waiting queue, bypassing newly arrived requests to maintain priority.

Default vLLM ignores both when input chunks arrive and when sequences are updated. A request that arrived early but has been idle receives higher priority than a request that arrived later but has freshly received input or updates, delaying progress on active requests. In update mode, DEFAULT vLLM prioritizes based on arrival time rather than recomputation cost. This causes requests with less recomputation work to be delayed behind requests requiring significant recomputation, extending their latency unnecessarily.

\subsubsection{FCFS (First-Come-First-Served)}
This scheduler separates requests into two tiers: \emph{full requests} (input sequence complete) scheduled in arrival order, and \emph{partial requests} (still receiving input) scheduled opportunistically. Eviction removes requests in reverse scheduling order when memory is exhausted.

The two-tier structure improves upon FIFO by separating complete from incomplete requests and prioritizing completed requests for output generation. However, FCFS shares FIFO's limitations: within each tier, scheduling uses only arrival order and ignores both chunk arrival recency and recomputation costs. Requests with older arrivals are prioritized over newer requests with fresh input or high-cost updates, delaying progress on active requests.

\subsubsection{MCPS (Most Chunks Processed Scheduling)}
Requests are prioritized by the number of tokens already computed (\texttt{num\_computed\_tokens}, highest first), with ties broken by arrival time. Eviction removes the request with the fewest computed tokens.

MCPS's progress-based metric works well in append mode, and naturally favors completing requests before starting new ones. Requests receiving chunks frequently maintain high priority and make steady progress. MCPS becomes problematic in update mode. When a request's input sequence is updated with a short LCP, \texttt{num\_computed\_tokens} resets to the LCP length (potentially near zero). A request that had computed many tokens suddenly drops to lowest priority, even though significant work has been done. This wastes prior computation and delays progress on requests undergoing frequent updates with short LCPs.

\subsubsection{LCAS (Last Chunk Arrival Scheduling)}
This scheduler combines two strategies: (1) separating complete and partial requests into two tiers, and (2) ordering both tiers by most recent chunk arrival time (\texttt{last\_chunk\_arrival\_time}, most recent first). Eviction removes the request with the oldest chunk arrival.

LCAS provides strong performance in append mode by prioritizing requests with recent chunk arrivals. When a chunk is appended at time $t$, the request is boosted to highest priority. The scheduler processes the expanded input sequence immediately while prior context is warm, maximizing efficiency. Requests transition from partial to complete tier as input finishes, and complete requests are prioritized for output generation. The approach naturally handles temporally clustered chunk arrivals well.

LCAS also performs well in update mode. 
When an input sequence is updated,
the request is boosted to high priority. This enables immediate recomputation of invalidated tokens, making progress on high-cost updates quickly. The priority boost benefits requests with short LCPs.
The main weakness in both modes is potential starvation: requests with infrequent chunk arrivals may be delayed indefinitely if other requests have more recent arrivals.

\section{Implementation and Usage}
\label{sec:implementation}

We implement \sys on top of the vLLM v1 engine, extending the core scheduler to support streaming inputs and the two-phase scheduling architecture described in Section~\ref{sec:scheduling-state-design}. The scheduler is modified to detect streaming request lifecycle events (initial request, input chunks, and completion) and invoke the KV cache manager to compute the longest common prefix (LCP) between old and new inputs, invalidating only the cache blocks corresponding to changed tokens. We implement the preemption decision framework as a cost comparison between recomputation and swapping strategies using performance models derived from hardware profiling. Multiple scheduling algorithms (Section~\ref{sec:scheduling-policies}) are implemented as priority sorting functions that can be selected at runtime via the \texttt{SCHEDULER\_TYPE} environment variable.

The KV cache manager is extended with GPU and CPU block pools to support both swapping and recomputation preemption strategies. When a request is preempted via swapping, its blocks are transferred to CPU memory and later swapped back when the request resumes. When preempted via recomputation, blocks are freed and the request recomputes affected tokens upon resumption. The evaluation drivers (crawler and ANNS) implement the streaming request lifecycle, submitting initial requests with the \texttt{is\_streaming\_prompt} flag, appending or updating inputs as chunks arrive (via \texttt{is\_prompt\_update}), and signaling completion with \texttt{is\_streaming\_prompt\_finished}. Event tracking records key state transitions (QUEUED, SCHEDULED, KV\_ON\_GPU, PREEMPTED\_SWAP, PREEMPTED\_RECOMPUTE, FINISHED) to enable detailed latency analysis and telemetry for streaming workloads.

\subsection{Public Interface}
\label{sec:interface}

\sys extends the vLLM engine interface with streaming-aware request objects and lifecycle management. The \texttt{EngineCoreRequest} object supports append and update modes via three key flags: \texttt{is\_streaming\_prompt} (incomplete input), \texttt{is\_streaming\_prompt\_finished} (signals completion), and \texttt{is\_prompt\_update} (toggles update vs. append). Clients may append input chunks or replace the input sequence entirely, and the engine automatically computes longest-common-prefix matches for cache invalidation.

The driver provides convenient functions for both modes: \texttt{new\_stream} (create initial request), \texttt{append} (add chunks), and \texttt{update} (replace input). Listing~\ref{lst:code-append} shows sample usage in both modes. 

\PP{Request Lifecycle}
Append mode submits an initial request with \texttt{is\_streaming\_prompt=True}, appends additional chunks as they arrive, then signals completion with \texttt{is\_streaming\_prompt\_finished=True}. Update mode follows the same pattern but sets \texttt{is\_prompt\_update=True} when replacing the entire input sequence.
The engine tracks both modes transparently and manages KV cache invalidation accordingly.

\begin{lstlisting}[style=PyStyle,caption={\small 
 Sample API usage in append and update modes.},label={lst:code-append}]
# e is engine object
# append mode
e.new_stream(rid="q1", ids=[query],...)
for chunk in chunks:
    e.append(rid="q1", ids=[chunk],...)
e.append(rid="q1", ..., finished=True)
# update mode
e.new_stream(rid="q2", ids=docs1+[q],...)
e.update(rid="q2", ids=docs2+[q],...)
e.update(rid="q2", ..., finished=True)
\end{lstlisting}

\PP{Output and Telemetry}
The engine returns \texttt{EngineCoreOutput} objects containing newly generated tokens, finish reason, and event timestamps. Events (QUEUED, SCHEDULED, KV\_ON\_GPU, PREEMPTED\_SWAP, PREEMPTED\_RECOMPUTE, FINISHED) enable latency analysis and scheduling diagnostics. The scheduler algorithm is selected via \texttt{SCHEDULER\_TYPE} environment variable (\texttt{DEFAULT\_VLLM}, \texttt{FCFS}, \texttt{MCPS}, or \texttt{LCAS}).

\section{Evaluation}
\label{sec:evaluation}

We evaluate \sys on two distinct streaming workloads: crawler-based context retrieval (append mode) and ANNS-based refinement (update mode). Our experiments demonstrate that scheduling policies designed for streaming inputs significantly improve responsiveness without sacrificing system throughput.

\begin{table}[t]
\centering
\footnotesize
\begin{tabular}{lcccc}
\toprule
\textbf{Metric} & \textbf{Mean} & \textbf{P50} & \textbf{P75} & \textbf{P95} \\
\midrule
\multicolumn{5}{l}{\textit{ANNS (Update Mode, 500 queries)}} \\
  Total Tokens per Query & 13K & 10K & 15K & 31K \\
  Retrieval Latency (s) & 4.5 & 3.9 & 5.3 & 8.5 \\
\midrule
\multicolumn{5}{l}{\textit{Crawler (Append Mode, 4,322 queries)}} \\
  Total Tokens per Query & 9.1K & 5.8K & 11.3K & 28.9K \\
  Retrieval Latency (s) & 9.9 & 9.3 & 11.6 & 16.7 \\
\bottomrule
\end{tabular}
\caption{Summary of streaming workload characteristics.}
\vspace{-1em}
\label{tab:workload-characteristics}
\end{table}
  
\subsection{Experiment Setup}
\minihead{Streaming Workloads} 
Due to the lack of large-scale, publicly available streaming workloads, we collected two realistic retriever traces with full response and timing information.
These traces are replayed at varying query-per-second (QPS) rates to emulate moderate load conditions where scheduling decisions are most impactful.
\autoref{tab:workload-characteristics} summarizes their characteristics.
  
\emph{\underline{Update Mode.}} We use the Fineweb-edu corpus (index: 372 GB, vectors: 279 GB) with SQuAD queries~\cite{rajpurkar-etal-2016-squad}, both encoded using \texttt{e5-base-v2}~\cite{wang2022text}.
We build a DiskANN index\cite{jayaram2019diskann} with L2 distance and configure search with beam width $W=8$ and list size $L=10000$.
This large $L$ increases retrieval accuracy but introduces significant latency, creating the high-latency, I/O-bound scenario \sys targets. 
We implement AquaPipe’s recall-aware prefetching~\cite{yu2025aquapipe}, which progressively refines the top-$k$ candidate list and enables early emission of partial results with sufficient recall.

\emph{\underline{Append Mode.}} We use the \verb|crawl4ai| Python library for core web crawling and scraping functionalities, which includes content pruning and link filtering with the BM25 algorithm to extract content only relevant to the original query~\cite{crawl4ai2024}.
Each crawl explores pages up to depth 2 and streams retrieved content to \sys in arrival order.
Queries are from OpenAI's SimpleQA dataset\cite{wei2024simpleqa}, which contains 4,327 fact-seeking questions requiring web-based evidence.
Our crawler workload applies per-document filtering and deduplication inline as each page is retrieved, enabling incremental streaming without a global reranking step.

\begin{figure}[t]
    \centering
    \includegraphics[width=\columnwidth, trim=0 0 0 35, clip]{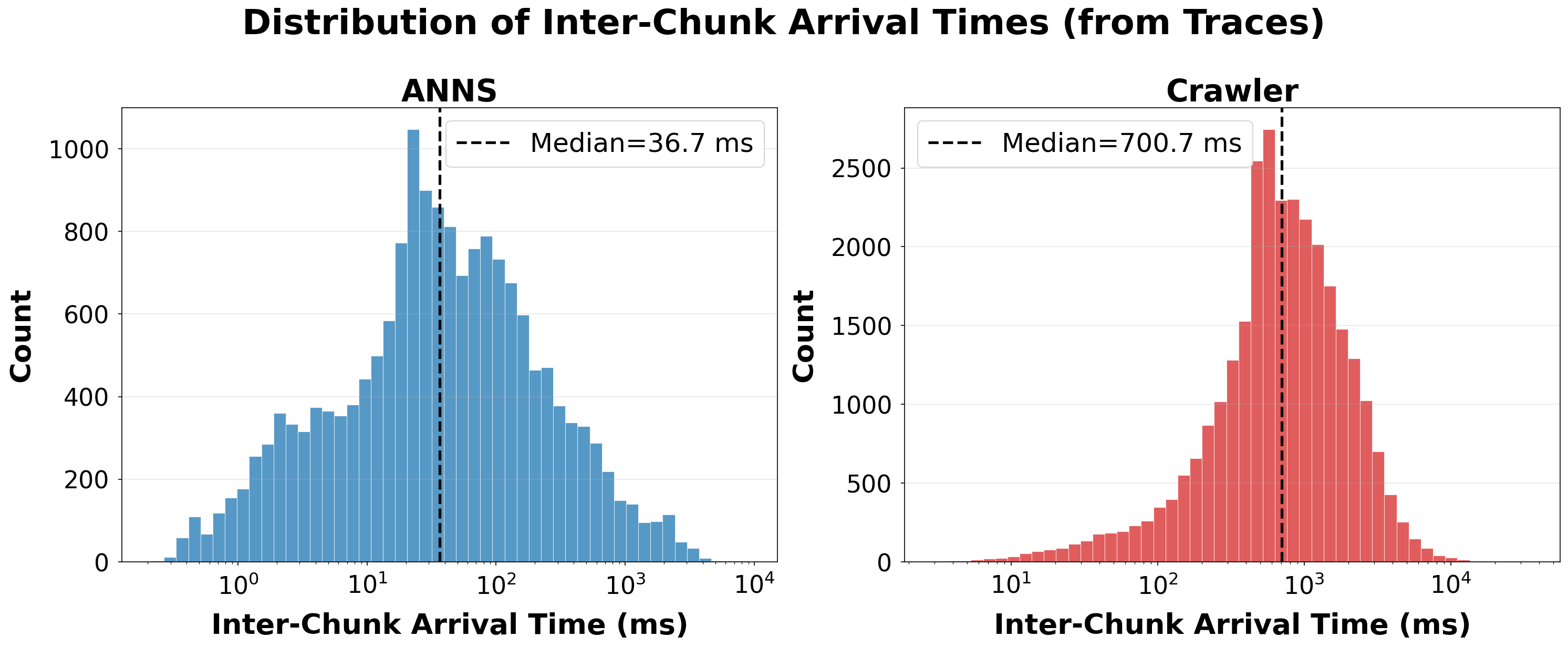}
    \vspace{-1em}
    \caption{Distribution of inter-chunk arrival times for ANNS and Crawler workloads (log-scale). ANNS chunks arrive with a median of 36.7\,ms, while crawler chunks arrive with a median of 700.7\,ms, exhibiting significantly higher variability.}
    \label{fig:chunk-arrival-histogram}
    \vspace{-1em}
\end{figure}

\begin{figure}[t]
\centering
\includegraphics[width=\columnwidth, trim=0 0 0 35, clip]{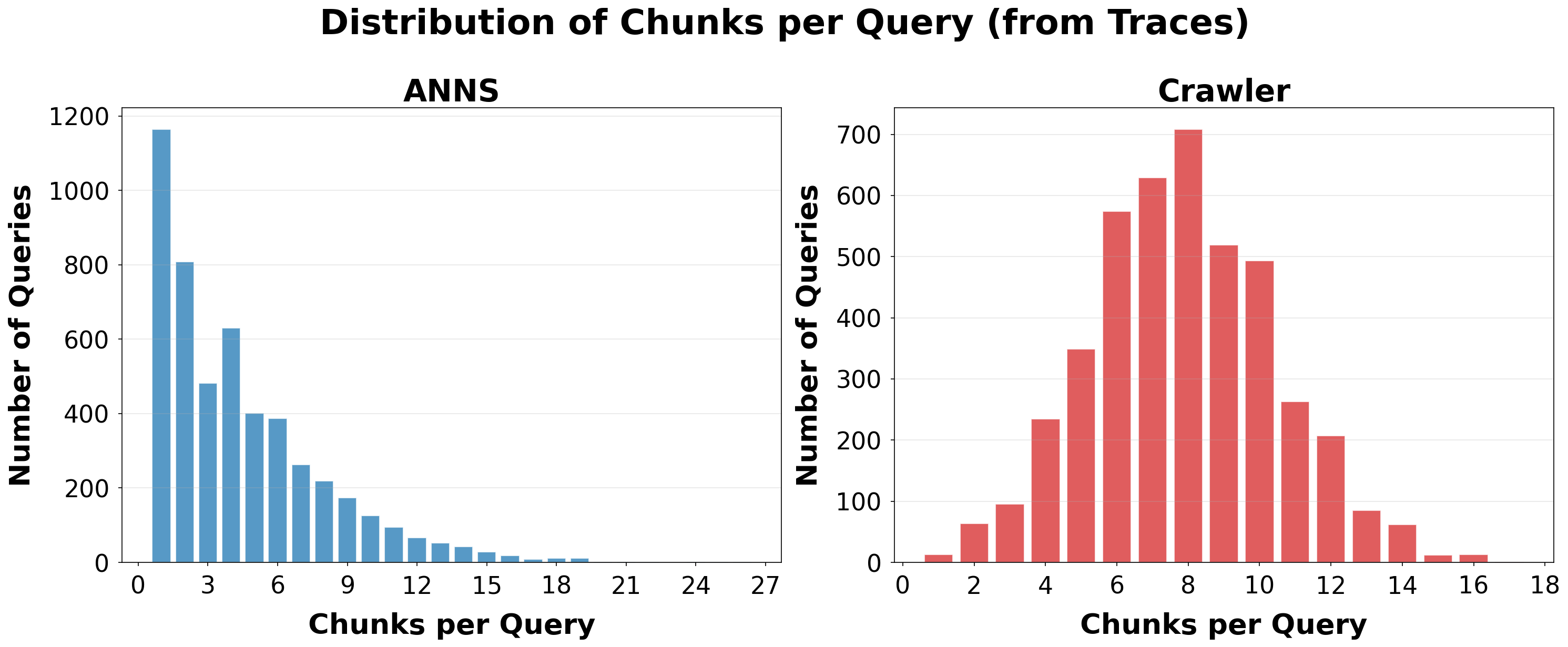}
\vspace{-1em}
\caption{Distribution of chunks per query. ANNS queries are heavily skewed toward 1--3 chunks, while crawler queries are concentrated around 6--10 chunks per query.}
\label{fig:chunks-per-query}
\vspace{-1em}
\end{figure}

\minihead{Chunk Arrival Patterns}
The two workloads exhibit fundamentally different chunk arrival patterns, which drive the scheduling challenges described in the main paper.

\autoref{fig:chunk-arrival-histogram} shows the distribution of inter-chunk arrival times.
ANNS arrivals are tightly concentrated around 36.7\,ms (median), with the bulk of the distribution spanning 1--1,000\,ms.
Crawler arrivals are 19$\times$ slower (median 700.7\,ms) and span approximately three orders of magnitude, from tens of milliseconds to over 30 seconds.
This high variability in crawler chunk arrivals creates extended idle periods for individual requests, motivating scheduling policies that can dynamically re-prioritize requests as new chunks arrive.

\autoref{fig:chunks-per-query} shows the distribution of chunks per query.
ANNS queries are heavily right-skewed, with the majority receiving 1--3 chunks ($>$50\% of queries receive $\leq$3 chunks).
Crawler queries follow a broader, roughly unimodal distribution centered around 6--10 chunks.
The combination of more chunks per query and higher per-chunk variability makes the crawler workload substantially more demanding for streaming schedulers.

\minihead{Hardware and Configuration}
Experiments run on NVIDIA H200 (141GB) and NVIDIA H100 (80GB) GPUs. We use Llama-3.1-8B-Instruct as the primary model. Tensor parallelism is set to 2, and GPU memory utilization target is 80\% (20\% buffer to prevent OOM). Token budget per scheduling step varies from 2048 to 8192. \sys extends capabilities of vLLM inference system.

\minihead{Methods of Comparison}
We compare \sys against two baselines:
(1) vLLM-NS: default vLLM scheduler without streaming support, and
(2) vLLM-S: default vLLM scheduler with streaming support.
We also evaluate \sys under its three custom scheduling and preemption policies--FCFS, MCPS, and LCAS--to assess the impact of scheduling strategy on latency and throughput.

\begin{figure*}[t]
  \centering
 \includegraphics[width=\textwidth]{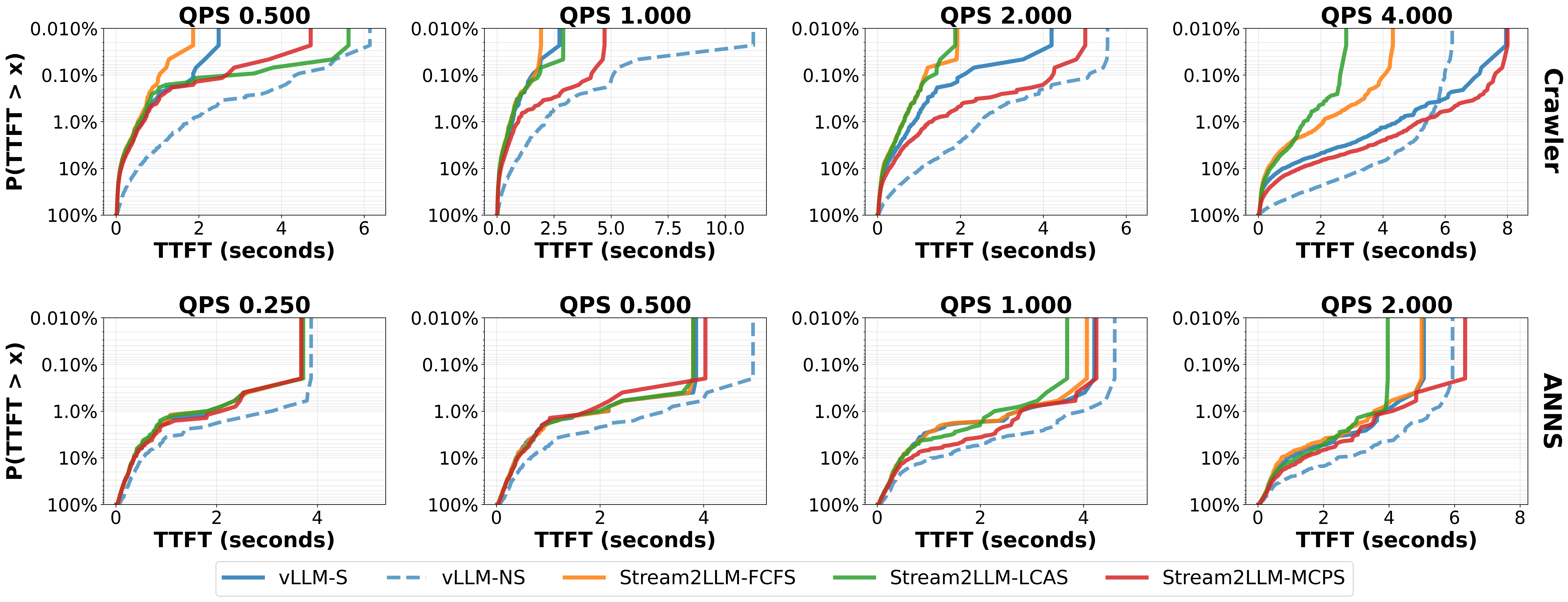}
  \vspace{-1em}
  \caption{TTFT CCDF across load levels for the crawler and ANNS workloads on H200. Streaming achieves up to 10.8–11.0$\times$ faster median latencies than non-streaming on the crawler workload, and up to 2.49–2.63$\times$ P95 speedups on the ANNS workload. }
  \vspace{-1.5em}
  \label{fig:ttft}
  \end{figure*}

\minihead{Metrics}
We report two key performance metrics: (1) \emph{TTFT (Time-To-First-Token)}, the time from request arrival to first generated token, which directly captures user-facing responsiveness, and (2) \emph{Trace Completion Time}, the total wall-clock time to complete all requests in a workload, reflecting system throughput and end-to-end performance.  
All experiments target the prefill instance in a disaggregated serving configuration; decode latency (TPOT) is handled by a separate decode instance and is not evaluated.

\subsection{Prefill Latency}
Figure~\ref{fig:ttft} shows TTFT performance on H200 across load levels for Crawler (top) and ANNS (bottom) workloads.

\textbf{Crawler Workload (Append Mode):} 
Streaming consistently improves TTFT across all loads. 
The performance gap between streaming and non-streaming widens with increasing load: at low loads (QPS 0.5–1.0), streaming achieves 3.9–4.3$\times$ faster median latencies over non-streaming, while at QPS 4.0, it delivers 10.8–11.0$\times$ faster median latencies by effectively overlapping retrieval and prefill. 

Within the streaming approaches, scheduler differentiation becomes apparent only when the page arrival rate approaches the prefill rate.
At QPS 2, FCFS's arrival-time ordering and LCAS's prioritization of recent page arrivals both achieve 1.44$\times$ and 1.32$\times$ P95 speedups over the default streaming baseline respectively, while MCPS's progress-based prioritization degrades to 0.73$\times$ as recent pages with fresh content are deprioritized.
This effect amplifies at QPS 4.0, where FCFS and LCAS reach 3.16$\times$ and 2.64$\times$ P95 speedups over the default streaming baseline, whereas MCPS falls to 0.71$\times$.
Beyond QPS 4.0, GPU utilization saturates and queuing delay dominates scheduling decisions, causing all approaches to converge to similar performance.

\begin{figure}[t]
\centering
\includegraphics[width=\columnwidth]{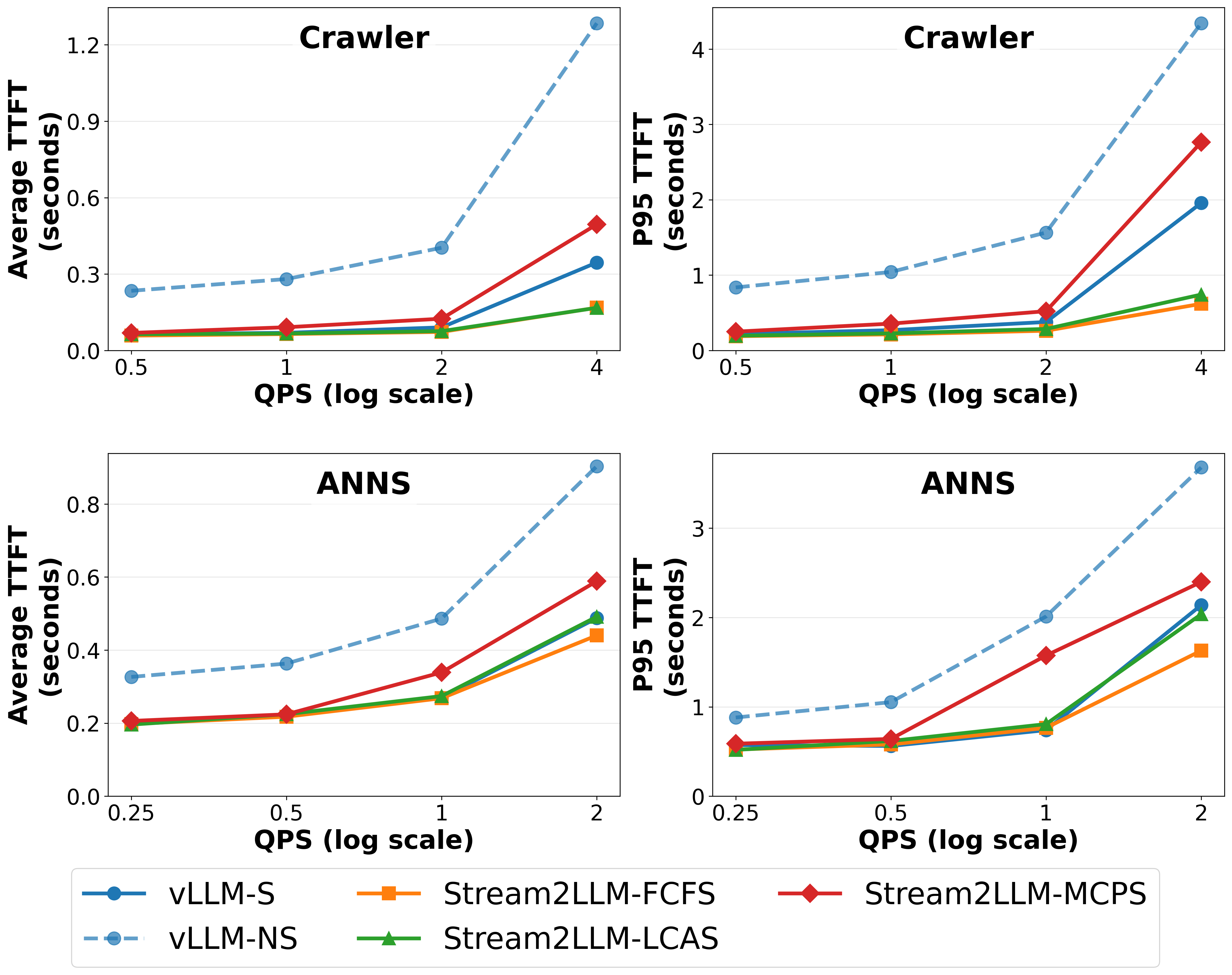}
\caption{Average and P95 time-to-first-token across schedulers at increasing request rates: crawler workload (top, 0.5–4 QPS) and ANNS (bottom, 0.25–2 QPS). }
\label{fig:ttft-qps}
\end{figure}

\begin{table*}[t]
\centering
\footnotesize
\caption{Scheduler and Eviction Strategy Ablation Study. Each cell shows TTFT speedup vs. Default vLLM non-streaming baseline (in secs) at different percentiles (P50/P99). \colorbox{red!15}{red} indicates degradation below the non-streaming baseline ($<$1$\times$). \colorbox{green!15}{green} indicates best speedup per column. Crawler: 4.0 QPS, 10$\times$ delays. ANNS: 2.0 QPS, 30$\times$ delays.}
\begin{tabular}{l|cc|cc|cc||cc|cc|cc}
\toprule
& \multicolumn{6}{c||}{\textbf{Crawler Workload}} & \multicolumn{6}{c}{\textbf{ANNS Workload}} \\
\cmidrule(lr){2-7} \cmidrule(lr){8-13}
& \multicolumn{2}{c|}{\textbf{Recompute}} & \multicolumn{2}{c|}{\textbf{Swap}} & \multicolumn{2}{c||}{\textbf{Cost-Based}} & \multicolumn{2}{c|}{\textbf{Recompute}} & \multicolumn{2}{c|}{\textbf{Swap}} & \multicolumn{2}{c}{\textbf{Cost-Based}} \\
\textbf{Scheduler} & P50 & P99 & P50 & P99 & P50 & P99 & P50 & P99 & P50 & P99 & P50 & P99 \\
\midrule
vLLM-NS   & 0.64s & 8.97s & 0.64s & 8.97s & 0.64s & 8.97s & 0.72s & 6.56s & 0.72s & 6.56s & 0.72s & 6.56s \\

vLLM-S    & 8.59$\times$ & \cellcolor{red!15}0.78$\times$ & 7.29$\times$ & \cellcolor{red!15}0.66$\times$ & 8.25$\times$ & \cellcolor{red!15}0.71$\times$ & 2.53$\times$ & \cellcolor{red!15}0.19$\times$ & 2.26$\times$ & \cellcolor{red!15}0.18$\times$ & 2.63$\times$ & \cellcolor{red!15}0.19$\times$ \\
FCFS                 & \cellcolor{green!15}8.77$\times$ & \cellcolor{green!15}10.03$\times$ & \cellcolor{green!15}7.78$\times$ & \cellcolor{green!15}6.69$\times$ & \cellcolor{green!15}8.30$\times$ & 8.62$\times$ & \cellcolor{green!15}2.70$\times$ & \cellcolor{green!15}1.84$\times$ & \cellcolor{green!15}2.37$\times$ & \cellcolor{green!15}1.26$\times$ & \cellcolor{green!15}2.70$\times$ & \cellcolor{green!15}2.04$\times$ \\
LCAS                 & 8.61$\times$ & 9.23$\times$ & 7.32$\times$ & 4.80$\times$ & 7.82$\times$ & \cellcolor{green!15}9.14$\times$ & 2.38$\times$ & 1.79$\times$ & 1.62$\times$ & \cellcolor{red!15}0.97$\times$ & 2.44$\times$ & 1.79$\times$ \\
MCPS                 & 5.92$\times$ & \cellcolor{red!15}0.73$\times$ & 3.86$\times$ & \cellcolor{red!15}0.48$\times$ & 4.96$\times$ & \cellcolor{red!15}0.77$\times$ & 2.20$\times$ & \cellcolor{red!15}0.19$\times$ & 1.47$\times$ & \cellcolor{red!15}0.22$\times$ & 2.23$\times$ & \cellcolor{red!15}0.19$\times$ \\
\bottomrule
\end{tabular}
\label{tab:eviction-ablation-combined}
\end{table*}

\begin{table}[t]
\centering
\footnotesize
\caption{Preemption statistics across workloads. Crawler (4.0 QPS, 10$\times$ delays) shows higher preemption with balanced swap/recompute. ANNS (2.0 QPS, 30$\times$ delays) shows lower frequency with cost-based heavily favoring recompute.}
\begin{tabular}{lrrr}
\toprule
\textbf{Policy} & \textbf{Recomp.} & \textbf{Swap} & \textbf{Cost-Based} \\
\midrule
\multicolumn{4}{c}{\emph{Crawler Workload}} \\
\cmidrule(lr){1-4}
vLLM-S & 2,448 & 779 & 1,735 (17\%/83\%) \\
FCFS   & 3,552 & 709 & 1,575 (21\%/79\%) \\
LCAS   & 12,564 & 1,049 & 3,486 (21\%/80\%) \\
MCPS   & 3,316 & 721 & 1,741 (19\%/81\%) \\
\midrule
\multicolumn{4}{c}{\emph{ANNS Workload}} \\
\cmidrule(lr){1-4}
vLLM-S & 130 & 35 & 125 (2\%/98\%) \\
FCFS   & 342 & 50 & 331 (1\%/99\%) \\
LCAS   & 385 & 48 & 373 (2\%/98\%) \\
MCPS   & 215 & 40 & 237 (0\%/100\%) \\
\bottomrule
\end{tabular}
\label{tab:preemption-stats-combined}
\end{table}

\textbf{ANNS Workload (Update Mode):}
Streaming also provides substantial and consistent benefits in update-mode workloads across all load levels. 
Similar to the crawler workload,  at low loads (QPS 0.25–0.5) and high loads ($>$QPS 2.0), all schedulers perform similarly--prefill rate dominates at low loads while queuing dominates at high loads.
At QPS 1.0, streaming schedulers converge tightly with 2.49--2.63$\times$ P95 speedups over non-streaming, indicating that prefill rate still dominates update arrival rate, and streaming architecture provides the dominant benefit regardless of scheduler choice. 
As load increases to QPS 2.0, scheduler differentiation emerges: FCFS achieves 2.26$\times$ P95 speedup over non-streaming, LCAS 1.81$\times$, and MCPS 1.53$\times$ over non-streaming.
MCPS slightly underperforms in this case because accumulated token progress becomes invalid when document sets change.
All maintain 1.30--1.42$\times$ P50 advantage over non-streaming. 
Streaming incurs significant cache invalidation
(\autoref{fig:tokens-invalidated-ccdf}): more than 10\% of requests at all loads invalidate over 10,000 tokens across all schedulers. Despite
this cost, streaming achieves up-to  $\sim 2\times$ faster TTFT than non-streaming, proving the responsiveness gains justify the trade-off. Beyond QPS 2.0, queueing delay dominates scheduling decisions, making cache efficiency gains secondary to responsiveness.

\subsection{Latency–Throughput Tradeoff}

\autoref{fig:ttft-qps} shows average and P95 TTFT as a function of QPS. As QPS increases, TTFT rises across all methods, but streaming schedulers consistently outperform the non-streaming baseline. In the crawler workload, FCFS and LCAS maintain the lowest TTFT at high QPS, while MCPS degrades. In the ANNS workload, all streaming schedulers track closely until QPS 2.0, where FCFS pulls ahead.

\autoref{fig:completion} (Appendix) confirms that streaming and scheduling optimizations do not affect aggregate throughput.
Trace completion times decrease hyperbolically with increasing QPS, and all scheduler variants (including non-streaming) produce indistinguishable curves on both workloads, demonstrating that the TTFT improvements come at no cost to system-level throughput.

\subsection{Performance under Memory Pressure}
\label{sec:mem-pressure}
Our default H200 setup (141GB memory, 80\% utilization) has sufficient memory that eliminates preemption. 
To evaluate scheduler behavior under memory pressure, we modified the workloads to saturate the GPU KV cache pool by increasing chunk delays: a factor of 10$\times$ for the crawler workload (append-mode) and 30$\times$ for ANNS (update-mode), where the delay multiplier is defined as $\frac{total\ KV\ tokens\ capacity}{avg\ tokens\ per\ query\times avg\ query\ duration \times QPS}$.

\textbf{Scheduler and Eviction Strategy Ablation:} We vary scheduler and eviction policy independently on both workloads.
\autoref{tab:eviction-ablation-combined} shows median (P50) and tail (P99) latency speedups versus the non-streaming baseline.
The study shows two key findings: (1) DEFAULT vLLM streaming exhibits catastrophic tail latency degradation under memory pressure---at P99, it performs 0.71$\times$ (crawler) and 0.19$\times$ (ANNS) vs.\ non-streaming baseline, demonstrating that streaming without proper scheduling is harmful under contention. (2) Eviction strategy choice creates substantial performance variation: in the crawler workload, FCFS achieves 10.03$\times$ speedup with recompute-only but only 6.69$\times$ with swap-only at P99, while LCAS shows similar sensitivity (9.23$\times$ vs.\ 4.80$\times$). The cost-based approach balances these extremes, achieving 8.62$\times$ (FCFS) and 9.14$\times$ (LCAS). For ANNS, the pattern repeats with smaller absolute speedups due to update-mode characteristics: FCFS achieves 1.84$\times$ (recompute) vs.\ 1.26$\times$ (swap) vs.\ 2.04$\times$ (cost-based) at P99. These results confirm that hardware-aware eviction is essential for performance under memory pressure, and proper scheduling (FCFS/LCAS) prevents catastrophic tail latency.

\textbf{Preemption Statistics:} \autoref{tab:preemption-stats-combined} reports preemption frequencies across configurations. The crawler workload triggers 779--12,564 total preemptions depending on scheduler aggressiveness (LCAS highest at 12.5K for recompute-only). Cost-based eviction allocates 17--21\% to SWAP and 79--83\% to RECOMPUTE for crawler, confirming balanced strategy selection. The ANNS workload exhibits lower preemption frequency (35--385 total) with cost-based heavily favoring RECOMPUTE (98--100\%), validating the earlier claim that update-mode characteristics make recomputation nearly always cost-optimal due to smaller effective KV caches after context invalidation.

\subsection{Overhead Analysis}

\minihead{Cache Efficiency in Update Mode} 
\autoref{fig:tokens-invalidated-ccdf} (Appendix) shows the CCDF of tokens invalidated per request across QPS levels for the ANNS workload.
All three streaming schedulers (FCFS, LCAS, MCPS) exhibit nearly overlapping invalidation curves, indicating that cache invalidation behavior is driven by the retrieval workload rather than the scheduling policy.
The non-streaming baseline (vLLM-NS) shows zero invalidation by design, as it waits for complete retrieval before beginning inference.
Invalidation patterns remain stable across load levels: at all QPS values, more than 10\% of requests invalidate over 10,000 tokens, reflecting the iterative document replacement in update mode.

\minihead{Scheduling Overhead} Scheduler sorting and budget allocation add sub-millisecond overhead per scheduling step. With 50 concurrent requests (representative of peak QPS), sorting latency is 15--16\,$\mu$s for FCFS/LCAS and 12--13\,$\mu$s for MCPS. Even at 500 requests, P99 latency remains below 165\,$\mu$s---negligible compared to prefill computation. The one-time offline profiling for hardware-specific cost models takes approximately five minutes per GPU configuration and is stored as a JSON file, imposing no runtime overhead.

\section{Related Work}
\label{sec:related}

\minihead{Retrieval–Inference Co-Design} Recent work has explored overlapping retrieval and inference to reduce end-to-end latency.
AquaPipe~\cite{yu2025aquapipe} improves time-to-first-token via recall-aware prefetching of intermediate ANNS results, but is limited to single-request settings and does not support scheduling with dynamic priorities.
It also relies on interrupting and restarting prefill when retrieval context changes, which is not supported in production systems such as vLLM.
PipeRAG~\cite{jiang2024piperag} uses pipeline parallelism with flexible retrieval intervals for iterative retrieval-generation, but evaluates with fixed batch sizes and does not address variable sequence lengths in deployments. 
Notably, PipeRAG targets encoder-decoder architectures with cross-attention, which is architecturally incompatible with the causal self-attention models used by \sys. 
Both systems leave open questions about scheduling policies for concurrency, adaptive preemption, and cache invalidation mechanisms across requests with different retrieval patterns. \sys addresses these by introducing a two-phase scheduling architecture supporting both append-mode and update-mode retrieval patterns in concurrent environments.

\minihead{LLM Inference Serving} vLLM~\cite{kwon2023efficient} provides PagedAttention for efficient KV cache storage and supports preemption via recomputation or swapping, but its scheduler assumes static input sequences. Orca~\cite{yu2022orca}, Sarathi~\cite{agrawal2024taming}, and FastServe~\cite{wu2023fast} improve batch efficiency and throughput through iteration-level scheduling, chunked prefills, and preemptive prioritization, respectively. 
Mooncake~\cite{qin2024mooncake}, DistServe~\cite{zhong2024distserve} and Splitwise~\cite{patel2023splitwise} use disaggregated GPU pools for prefill and decode. These systems assume static input sequences and do not support dynamic updates or cache invalidation when sequences change dynamically.

\minihead{Scheduling and Cache Optimization} Scheduling policies range from FCFS to SJF variants~\cite{patel2023splitwise} and sophisticated approaches like Sarathi-Serve~\cite{agrawal2024taming} that minimize pipeline stalls. However, existing policies assume static input sequences and do not account for dynamic priority shifts from streaming context arrival. For cache optimization, PagedAttention divides KV cache into fixed blocks, while RadixAttention~\cite{zheng2024radix} enables fine-grained sharing via radix trees. These optimize memory for static input sequences but lack mechanisms for cache invalidation when sequences change. \sys addresses this with longest common prefix (LCP)-based cache invalidation that preserves valid blocks while selectively invalidating only affected blocks, minimizing recomputation across dynamic retrieval patterns.

\section{Conclusion}
\label{sec:conclusion}
\sys bridges the latency gap in context retrieval systems by enabling streaming input processing with concurrent requests. Its decoupled scheduling architecture, LCP-based cache invalidation, and hardware-aware preemption models extend vLLM to handle dynamically evolving contexts efficiently. Our evaluation shows that streaming delivers order-of-magnitude improvements in time-to-first-token latency without sacrificing throughput. Scheduler choice has limited impact when memory is abundant but becomes critical under pressure, improving append-mode performance and preventing degradation in update-mode workloads. \sys establishes streaming as a core architectural primitive for low-latency, high-throughput context retrieval in large-scale LLM deployments.

\section*{Acknowledgments}
This work was supported in part by the National Science Foundation under grant IIS-2335881.

\bibliography{main}
\bibliographystyle{main}

%%%%%%%%%%%%%%%%%%%%%%%%%%%%%%%%%%%%%%%%%%%%%%%%%%%%%%%%%%%%%%%%%%%%%%%%%%%%%%%
% SUPPLEMENTAL CONTENT AS APPENDIX AFTER REFERENCES
%%%%%%%%%%%%%%%%%%%%%%%%%%%%%%%%%%%%%%%%%%%%%%%%%%%%%%%%%%%%%%%%%%%%%%%%%%%%%%%
\clearpage

\appendix
\newpage
\section{Additional Evaluation}
\label{sec:appendix-eval}

\subsection{Cache Invalidation in Update Mode}

\begin{figure*}[t]
\centering
\includegraphics[width=\textwidth]{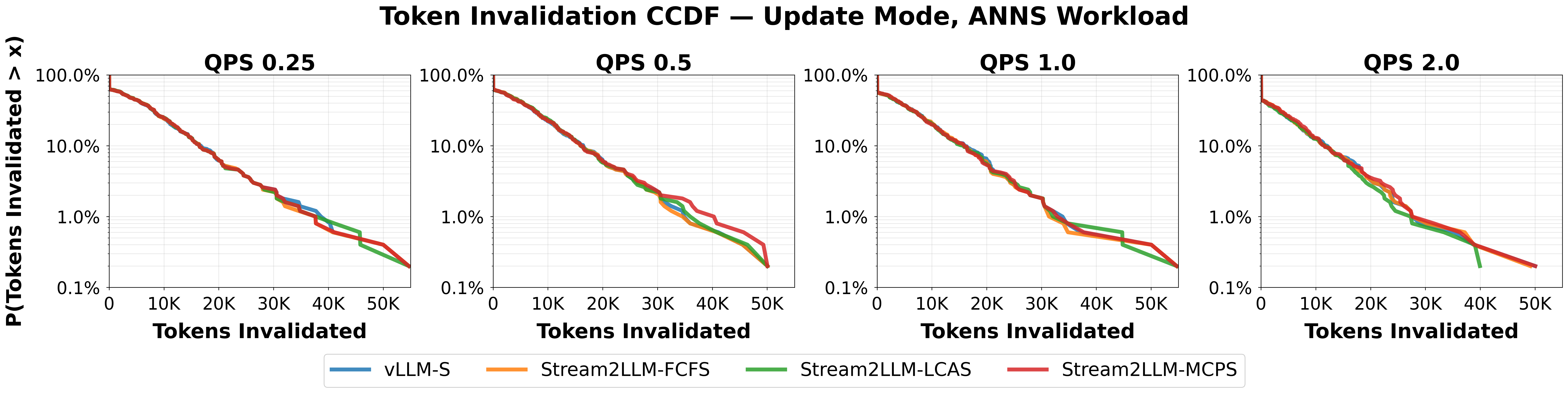}
\vspace{-1em}
\caption{Tokens invalidated per request (CCDF) across QPS levels (0.25--2.0) for the ANNS workload. All \sys schedulers show similar cache invalidation behavior. vLLM-NS has zero invalidation as it waits for complete retrieval.}
\vspace{-1em}
\label{fig:tokens-invalidated-ccdf}
\end{figure*}

\autoref{fig:tokens-invalidated-ccdf} shows the CCDF of tokens invalidated per request across QPS levels for the ANNS workload.
All three \sys schedulers (\sys-FCFS, \sys-LCAS, \sys-MCPS) exhibit nearly overlapping invalidation curves, indicating that cache invalidation behavior is driven by the retrieval workload rather than the scheduling policy.
The non-streaming baseline (vLLM-NS) shows zero invalidation by design, as it waits for complete retrieval before beginning inference.
Invalidation patterns remain stable across load levels: at all QPS values, more than 10\% of requests invalidate over 10,000 tokens, reflecting the update-mode characteristic of iterative document replacement.

\begin{figure}[t]
\centering
\includegraphics[width=\columnwidth]{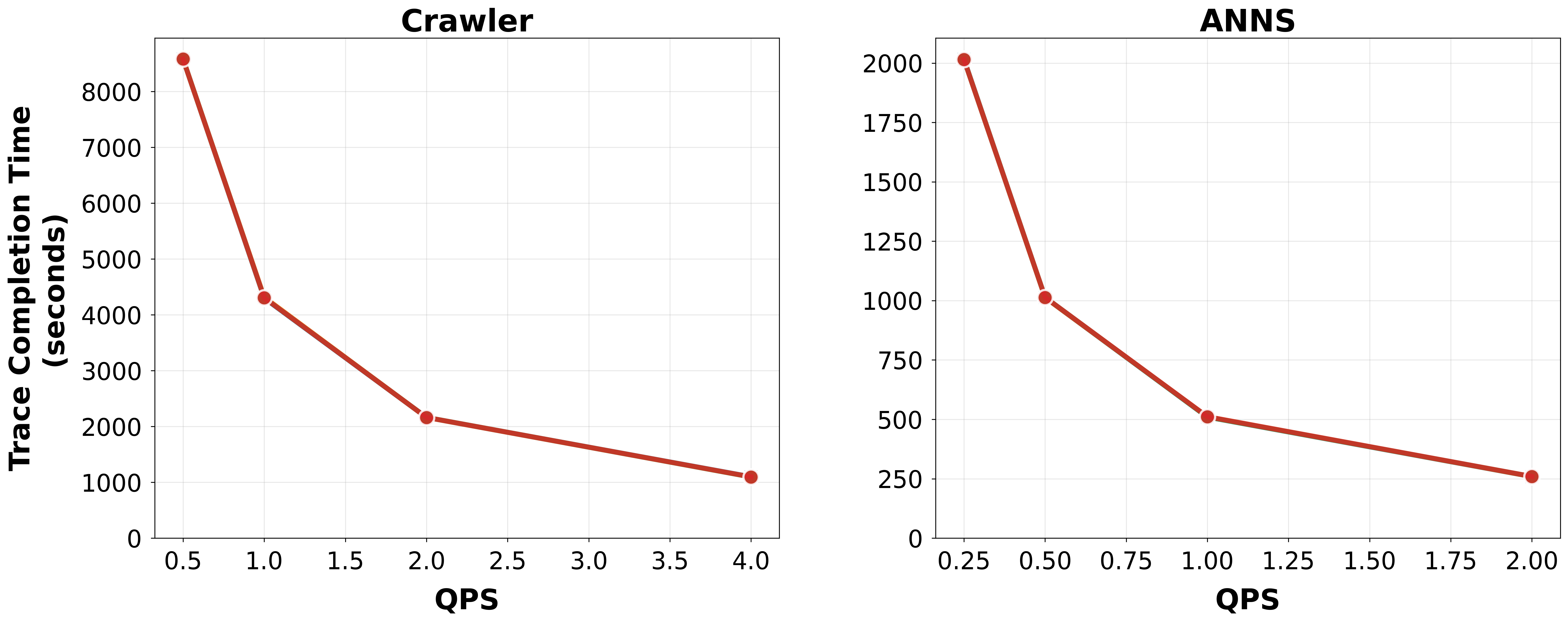}
\vspace{-1em}
\caption{Trace completion time across QPS levels for both workloads. All scheduler variants achieve near-identical completion times, confirming throughput parity.}
\label{fig:completion}
\vspace{-1em}
\end{figure}

\subsection{Throughput Parity}

\autoref{fig:completion} confirms that streaming and scheduling optimizations do not affect aggregate throughput.
Trace completion times decrease hyperbolically with increasing QPS, and all scheduler variants (including non-streaming) produce indistinguishable curves---appearing as a single overlapping line---on both workloads.
This demonstrates that the TTFT improvements reported in the main evaluation come at no cost to system-level throughput.

\section{Artifact Appendix}

\subsection{Abstract}

This artifact contains the open-source Stream2LLM system: a modified vLLM~0.8.1 engine with streaming input support, along with all scripts, pre-computed run logs, and workload traces needed to reproduce every figure, table, and inline number in the paper.
The repository includes two evaluation paths: (1)~an \emph{artifact-only} path that regenerates all paper artifacts from pre-computed data in approximately five minutes without GPU hardware, and (2)~a \emph{full re-run} path that re-executes the experiments end-to-end on NVIDIA GPUs.
For a detailed version of this artifact appendix, see the repository \texttt{README.md}.

\subsection{Artifact check-list (meta-information)}

{\small
\begin{itemize}
  \item {\bf Algorithm:} vLLM-NS, vLLM-S, \sys-FCFS, \sys-LCAS, \sys-MCPS scheduling; cost-based preemption
  \item {\bf Program:} Python scripts; modified vLLM~0.8.1 engine
  \item {\bf Data set:} Web-crawler traces; ANNS pipeline traces; performance-model JSONs
  \item {\bf Run-time environment:} Linux, Python~3.10.9, CUDA, conda, pip
  \item {\bf Hardware:} Artifact-only: any machine. Full re-run: 2$\times$ H200 or 2$\times$ H100 GPUs
  \item {\bf Metrics:} TTFT, trace completion time, preemption counts, scheduler sorting latency
  \item {\bf Output:} Figures in \texttt{figures/}, tables in \texttt{tables/}, logs in \texttt{data/run\_log/}
  \item {\bf How much disk space required (approximately)?:} ${\sim}$5\,GB
  \item {\bf How much time is needed to prepare workflow (approximately)?:} ${\sim}$15 minutes
  \item {\bf How much time is needed to complete experiments (approximately)?:} ${\sim}$5 minutes (artifact-only); ${\sim}$48 hours (full re-run)
  \item {\bf Publicly available?:} Yes
  \item {\bf Code licenses (if publicly available)?:} MIT
  \item {\bf Data licenses (if publicly available)?:} CC-BY-4.0
  \item {\bf Archived (provide DOI)?:} \url{https://doi.org/10.5281/zenodo.18906769}
\end{itemize}
}

\subsection{Description}

\subsubsection{How delivered}

The artifact is delivered as a GitHub repository on the \texttt{mlsys\_artifact} branch:

\begin{center}
\url{https://github.com/rajveerb/stream2llm}
\end{center}

\noindent All large data (run logs, workload traces, performance-model JSONs) is stored in a git submodule hosted on HuggingFace:

\begin{center}
\url{https://huggingface.co/datasets/rbachkaniwala3/stream2llm-data}
\end{center}

\subsubsection{Hardware dependencies}

No GPU is required for artifact-only evaluation.
For full experiment re-runs, the paper's experiments used NVIDIA 2$\times$H200 and 2$\times$H100 GPUs.
See \texttt{README.md} for details.

\subsubsection{Software dependencies}

Dependencies are installed in a \texttt{conda} environment via \texttt{pip}.
See \texttt{README.md} for setup instructions and \texttt{requirements.txt} for the full package list.

\subsubsection{Data sets}

All data is hosted as a HuggingFace dataset (\texttt{rbachkaniwala3/stream2llm-data}) and fetched automatically via the git submodule. See the ``Data Organization'' section of \texttt{README.md} for details.

\subsection{Installation}

Refer to the \texttt{README.md}.

\subsection{Experiment workflow}

\subsubsection{Artifact-only (no GPU required)}

\begin{verbatim}
bash reproduce_artifacts.sh
\end{verbatim}

\noindent This generates all figures in \texttt{figures/} and analysis tables in \texttt{tables/} from pre-computed run logs. Takes approximately five minutes.

\subsubsection{Full re-run (GPU required)}

The experiment drivers run each scheduler variant (\texttt{default\_vllm}, \texttt{fcfs}, \texttt{lcas}, \texttt{mcps}) across all arrival-time configurations and write logs to \texttt{data/run\_log/}.
See \texttt{README.md} (``Building Stream2LLM Engine'' and ``Running Experiments'') and \texttt{experiments/README.md} for the full list of experiment configurations and commands.

\subsection{Evaluation and expected result}
\textbf{Figures and tables.} Generated figures in \texttt{figures/} should visually match the pre-built reference copies in \texttt{figures/reference/}. Analysis scripts produce \texttt{.txt} files in \texttt{tables/}. The mapping from paper artifacts to scripts is:

  {\small
  \begin{itemize}
    \item \Cref{fig:perf-model} $\to$ \texttt{plot\_recomp\_vs\_swap\_clean.py}
    \item \Cref{fig:chunk-arrival-histogram,fig:chunks-per-query} $\to$ \texttt{chunk\_arrival\_characterization.py}
    \item \Cref{fig:ttft} $\to$ \texttt{plot\_ttft\_ccdf\_stacked\_2x4.py}
    \item \Cref{fig:ttft-qps} $\to$ \texttt{crawler/plotter\_utils/plot\_ttft\_qps\_comparison.py} (Crawler half) and \texttt{anns/plotter\_utils/plot\_ttft\_qps\_comparison.py} (ANNS half)
    \item \Cref{fig:completion} $\to$ \texttt{plot\_trace\_completion\_combined.py}
    \item \Cref{fig:tokens-invalidated-ccdf} $\to$ \texttt{plot\_tokens\_invalidated\_aggregated.py}

    \item \Cref{tab:eviction-ablation-combined} $\to$ \texttt{compute\_scheduler\_improvements.py}
    \item \Cref{tab:preemption-stats-combined} $\to$ \texttt{analyze\_preemptions.py}
    \item Scheduler latency $\to$ \texttt{benchmark\_scheduler\_latency.py}
  \end{itemize}
  }

\subsection{Experiment customization}

\begin{itemize}
  \item \textbf{Scheduler selection:} per-scheduler YAML configs in \texttt{experiments/\{crawler,anns\}/configs/}
  \item \textbf{Arrival-time sweeps:} supported via the driver script's \texttt{scheduler} and \texttt{compare} modes
  \item \textbf{Hardware adaptation:} modify YAML configs and performance-model JSONs in \texttt{data/perf\_model/}
\end{itemize}

\noindent See \texttt{experiments/README.md} for full details.

\subsection{Methodology}

Submission, reviewing, and badging methodology:

\begin{itemize}
  \item \url{http://cTuning.org/ae/submission-20190109.html}
  \item \url{http://cTuning.org/ae/reviewing-20190109.html}
  \item \url{https://www.acm.org/publications/policies/artifact-review-badging}
\end{itemize}

\end{document}